\documentclass{article}
\usepackage[a4paper,margin=1in]{geometry}
\usepackage{graphicx}
\usepackage{xcolor}
\usepackage{standalone}
\usepackage{tikz}
\usepackage{hyperref,url}
\hypersetup{
    colorlinks,
    linkcolor=black,
    citecolor=blue,
    urlcolor={blue!80!black}
}

\usepackage{amsmath,amssymb,dsfont}
\usepackage{pifont}
\newcommand{\cmark}{\ding{51}}
\newcommand{\xmark}{\ding{55}}
\usepackage{multirow,array}
\usepackage{authblk}
\usepackage{changepage}

\usepackage[round]{natbib}
\AtBeginDocument{\def\doi#1{\url{https://doi.org/#1}}}

\usepackage{subcaption}
\usepackage{threeparttable}
\usepackage{array,multirow}
\usepackage{makecell}
\newcolumntype{L}[1]{>{\raggedright\let\newline\\\arraybackslash\hspace{0pt}}m{#1}}
\newcolumntype{C}[1]{>{\centering\let\newline\\\arraybackslash\hspace{0pt}}m{#1}}
\newcolumntype{R}[1]{>{\raggedleft\let\newline\\\arraybackslash\hspace{0pt}}m{#1}}
\newcolumntype{P}[1]{>{\centering\arraybackslash}p{#1}}
\newcolumntype{M}[1]{>{\centering\arraybackslash}m{#1}}

\newcommand{\rmd}{\mathrm{d}}
\newcommand{\bbE}{\mathbb{E}}
\newcommand{\bbR}{\mathbb{R}}

\newcommand{\Var}{\mathbb{V}\mathrm{ar}}

\newcommand{\CRPS}{\mathrm{CRPS}}
\newcommand{\rmS}{\mathrm{S}}

\title{Distributional Regression U-Nets for the Postprocessing of Precipitation Ensemble Forecasts}
\author[1]{Romain Pic}
\author[1]{Clément Dombry}
\author[2]{Philippe Naveau}
\author[3]{Maxime Taillardat}

\affil[1]{\small Université de Franche Comté, CNRS, LmB (UMR 6623), F-25000 Besançon, France}
\affil[2]{Laboratoire des Sciences du Climat et de l'Environnement, UMR 8212, CEA-CNRS-UVSQ, EstimR, IPSL \& U Paris-Saclay, Gif-sur-Yvette, France}
\affil[3]{CNRM, Université de Toulouse, Météo-France, CNRS, Toulouse, France}

\begin{document}

\maketitle

\begin{abstract}
    Accurate precipitation forecasts have a high socio-economic value due to their role in decision-making in various fields such as transport networks and farming. We propose a global statistical postprocessing method for grid-based precipitation ensemble forecasts. This U-Net-based distributional regression method predicts marginal distributions in the form of parametric distributions inferred by scoring rule minimization. Distributional regression U-Nets are compared to state-of-the-art postprocessing methods for daily 21-h forecasts of 3-h accumulated precipitation over the South of France. Training data comes from the Météo-France weather model AROME-EPS and spans 3 years. A practical challenge appears when consistent data or reforecasts are not available.

    Distributional regression U-Nets compete favorably with the raw ensemble. In terms of continuous ranked probability score, they reach a performance comparable to quantile regression forests (QRF). However, they are unable to provide calibrated forecasts in areas associated with high climatological precipitation. In terms of predictive power for heavy precipitation events, they outperform both QRF and semi-parametric QRF with tail extensions.
\end{abstract}

\section{Introduction}

Correctly forecasting precipitation is crucial for decision-making in various fields such as flood levels, transport networks, water resources and farming, among others (see, e.g., \citealt{Olson1995}). Moreover, high-impact events are expected to intensify in the future as a consequence of climate change \citep{Planton2008}. Numerical weather prediction (NWP) systems have been continuously improving to take into account uncertainty of the atmosphere and the limitations of their physical modeling \citep{Bauer2015}. NWP systems produce ensemble forecasts, consisting of multiple runs of deterministic scenarios with different parameters. Nonetheless, raw ensemble forecasts suffer from bias and underdispersion (see, e.g., \citealt{Hamill1997, Bauer2015, BenBouallegue2016, Baran2016}). This phenomenon affects all NWP systems regardless of the weather service and of the variable of interest. Furthermore, the limited number of ensemble members coupled with underdispersion implies that raw ensemble forecasts may have a limited predictive power regarding extremes \citep{Williams2013}. In order to correct these systematic errors, it has become standard practice to use statistical postprocessing of ensemble prediction systems (EPS) in both research and operations.\\

A popular spatial statistical postprocessing strategy consists of separately postprocessing marginal distributions at each location and the spatial dependence structure. Numerous methods for postprocessing univariate marginals have been developed over the past two decades. There has been a rise in the number of machine learning based statistical postprocessing techniques as they provide a flexible framework enabling the modeling of complex relationships between the output of NWP models and the target variable. Moreover, they facilitate the use of a large number of predictors. These methods range from well-established statistical learning techniques, such as random forests \citep{Taillardat2016} or gradient boosting \citep{Messner2017}, to neural networks or deep learning techniques, such as fully connected neural networks \citep{Rasp2018} and transformers \citep{BenBouallegue2024}. For a thorough review of the existing statistical postprocessing techniques, readers may refer to \cite{Vannitsem2021} and \cite{Schulz2022}.
Once calibrated univariate marginals are obtained, the spatial dependence structure may be needed by downstream applications. The spatial dependence structure can be obtained from the raw ensemble as done by ensemble copula coupling (ECC; \citealt{Schefzik2013}) and its variants (e.g., \citealt{BenBouallegue2016}) or from historical observations as done by Schaake shuffle (ScS; \citealt{Clark2004}). Alternatively, if raw ensembles or historical data do not model the spatial dependence sufficiently well, it can be postprocessed using adapted techniques (see, e.g., \citealt{Schefzik2018}).

An alternative postprocessing strategy consists of direct postprocessing of raw ensemble members to obtain calibrated members. This can be achieved by postprocessing each member individually \citep{VanSchaeybroeck2014} or by using ensemble-agnostic methods \citep{BenBouallegue2024}.\\

In order to circumvent (potential) data scarcity, it is common to use parametric methods as they are usually less affected by smaller training datasets. The choice of a specific parametric distribution can be motivated by prior knowledge (or assumption) on the distribution of the variable of interest. Parametric methods can enable extrapolation beyond the range available in the training data, which is of interest to consider extreme events (see, e.g., \citealt{Friederichs2018} and \citealt{Taillardat2019}). In particular, certain meteorological variables have a heavy-tailed distribution; thus, a parametric method can be used to ensure that postprocessed distributions will have an appropriate tail behavior (e.g., \citealt{Lerch2013}).

Previous studies, such as \cite{Hemri2014} and \cite{Taillardat2020}, have highlighted that all meteorological quantities do not represent the same difficulty in terms of postprocessing. Variables with heavy-tailed climatological distributions or variables with short-scale spatio-temporal dependence (e.g., rainfall or wind gusts) are more difficult to treat than light-tailed variables or spatially smooth variables (e.g., surface temperature or sea level pressure). In the same vein, \cite{Schulz2022} states that "wind gusts are a challenging meteorological target variable as they are driven by small-scale processes and local occurrence, so that their predictability is limited even for numerical weather prediction (NWP) models run at convection-permitting resolutions."\\

NWP models produce forecasts on a grid that are of interest to downstream applications \citep[Section~7.3.2]{Hamill2018}. However, consistent gridded data suited to postprocessing is computationally costly since reanalyses and reforecasts of gridded products are demanding in terms of both storage and computation. Numerous observation networks are station-based (e.g., temperature, wind speed, or pressure), but they vary in coverage and quality. When forecasts are required at nearby locations, spatial modeling procedures are required. Both station-based and grid-based approaches present benefits and drawbacks \citep[Section~7.3.2]{Hamill2018}. No preference has reached a consensus for any variable, but \cite{Feldmann2019} shows that the relative improvement is greater for station-based 2-m temperature postprocessing when station-based observations are used. In the case of precipitation, observations can be measured by hybrid observations (gauge-adjusted radar images), allowing for improvement in the quality of gridded postprocessing.

As mentioned by \cite{Schulz2022}, one of the main challenges of postprocessing is to preserve the spatio-temporal information while optimally utilizing the whole available input data. This motivates the use of global statistical postprocessing models (e.g., a single model for multiple locations). Distributional regression networks (DRN; \citealt{Rasp2018}) use an embedding module to learn a representation of stations, allowing the model to learn from nearby and similar stations in order to preserve the spatial information of the data. When working with gridded data, a postprocessing method could benefit from taking into account this spatial structure of the data within its architecture. Convolutional neural networks (CNN) rely on the image-like structure of their input. Numerous CNN-based methods have been developed to perform postprocessing (see, e.g., \citealt{Dai2021} and \citealt{Lerch2022}). Here, we want the output of the statistical postprocessing method to be grid-based. U-Net \citep{Ronneberger2015} architectures appear to be a natural solution to preserve the spatial structure of the data. U-Nets use a sequence of convolutional blocks to learn complex features and upscaling blocks to retrieve parameters of interest at the desired resolution. We propose a U-Net-based method to postprocess marginals at each grid point using predictors at nearby grid points for high-resolution precipitation ensemble forecasts.\\

The paper is organized as follows. Section~\ref{section:data} presents the dataset used in this study. In Section~\ref{section:methods}, three state-of-the-art methods composing the reference methods of this study, namely quantile regression forests (QRF), QRF with tail extension (TQRF) and DRN, are presented and compared based on their known benefits and limitations. A U-Net-based method, called distributional regression U-Nets (DRU), is introduced and compared with U-Net-based postprocessing methods in the literature. The predictive performance of the models is compared in terms of multiple univariate metrics in Section~\ref{section:results}. An emphasis is put on the predictive performance of extremes. Finally, Section~\ref{section:discussion} sums up the performance of DRU and offers possible perspectives.\\

The code used to implement the different methods and their verification is publicly available\footnote{\url{https://github.com/pic-romain/unet-pp}}.

\section{Data}\label{section:data}

\begin{figure}
    \centering
    \includegraphics[width=15cm]{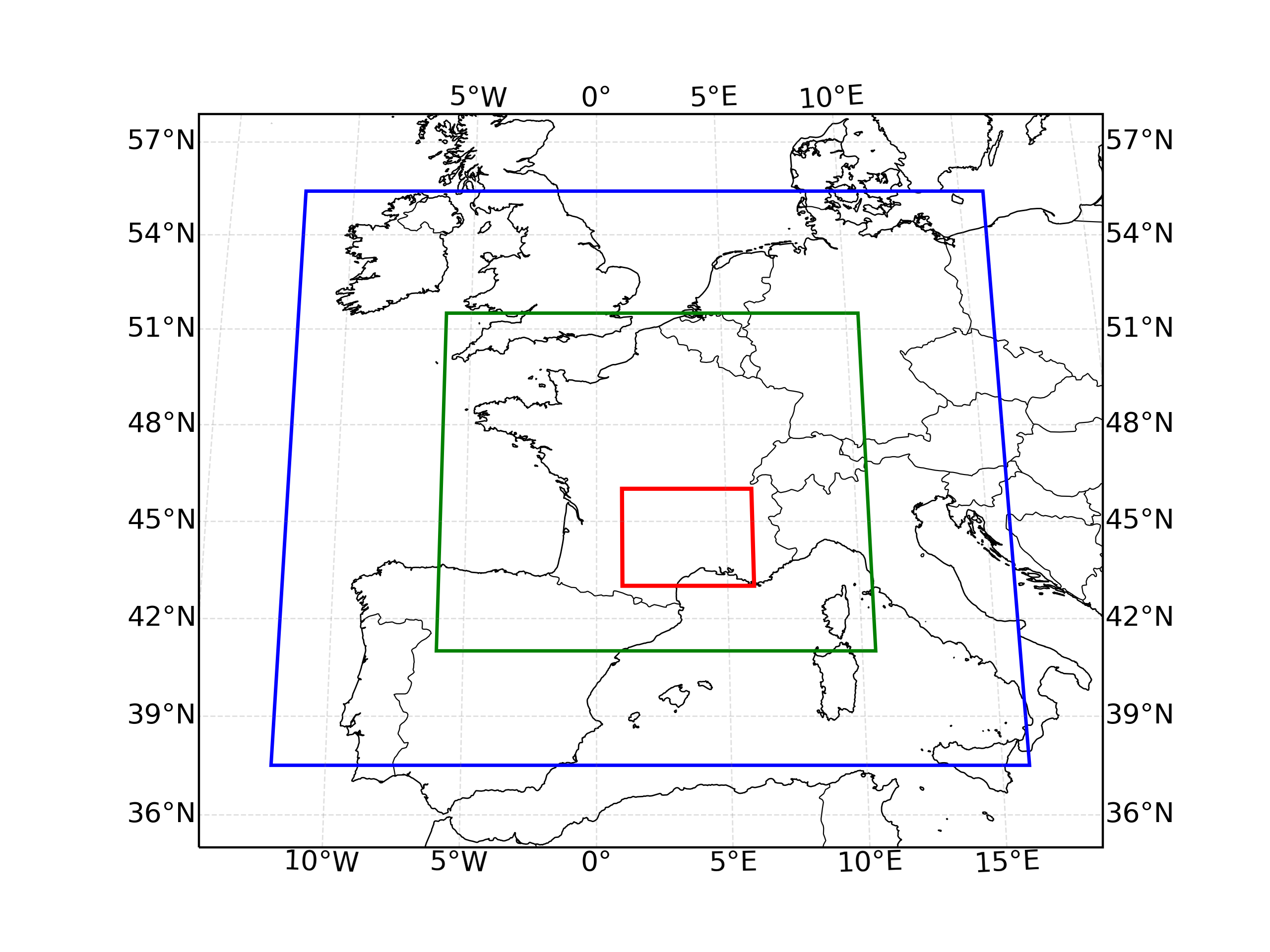}
    \caption{Domains covered by AROME-EPS (blue), ANTILOPE (green) and the region of interest (red).}
    \label{fig:domains}
\end{figure}

In this study, we focus on 3-h accumulated precipitation over the South of France (see Fig.~\ref{fig:domains}) at a forecast lead time of 21-h initialized at 15:00UTC daily. Ensemble forecasts are taken from the 17-member limited area ensemble forecasting system AROME-EPS \citep{Bouttier2015} driven by a subsampling of the global\footnote{in the sense of globe-wide} PEARP ensemble. AROME-EPS produces ensembles with one control member and 16 perturbed members for forecasts up to 51 hours on four different initialization times. It produces a gridded ensemble over Western Europe with a horizontal resolution of 0.025$^{\circ}$ based on a model run at 1.3 km resolution. The probabilistic forecasts are compared to 3-h accumulated precipitation data obtained from the gauge-adjusted radar product ANTILOPE \citep{Champeaux2009}, which has a spatial resolution of 0.001$^{\circ}$ over Western Europe. We project observations of ANTILOPE onto the AROME-EPS grid using bilinear interpolation.\\

The region of interest in this study covers areas, such as the \textit{Cévennes}, prone to heavy precipitation events (HPEs) \citep{Ricard2012}. HPEs affect Mediterranean coastal regions regularly causing flash floods. Mediterranean HPEs are typically characterized by quasi-stationary convective precipitation and may have limited predictability due to their intensity and being very local \citep{Caumont2021}. Statistical postprocessing methods can help improve forecasting such events.\\

Our period of interest spans 4 years from November 2019 to October 2023. The period from November 2019 to October 2022 is used as a training/validation dataset using 7-fold cross-validation to tune hyperparameters of the models. The folds are based on the day of the week. The period from November 2022 to October 2023 is used as a hold-out test set. All the results of Section~\ref{section:results} are provided for models trained on the entirety of the training/validation dataset and evaluated on the test dataset. The dataset is composed of forecasts and reforecasts from two different cycles of AROME-EPS. Consistency of both raw ensembles and observations is important since independent and identically distributed (i.i.d.) data is assumed. The two cycles of AROME-EPS used, namely 43t2 and 46t1, only have minor differences, making the i.i.d. assumption reasonable.\\

We use summary statistics of the AROME-EPS ensemble as predictors. The following variables were selected based on experts' opinions: precipitation, convective available potential energy, maximal reflectivity, pseudo wet-bulb potential temperature, relative humidity and AROME convection index. For each of these variables, the mean, the minimum, the maximum and the standard deviation of the raw ensemble were computed at each grid point and used as predictors.\\

\begin{figure}[!ht]
    \centering
    \includegraphics[width=.95\textwidth]{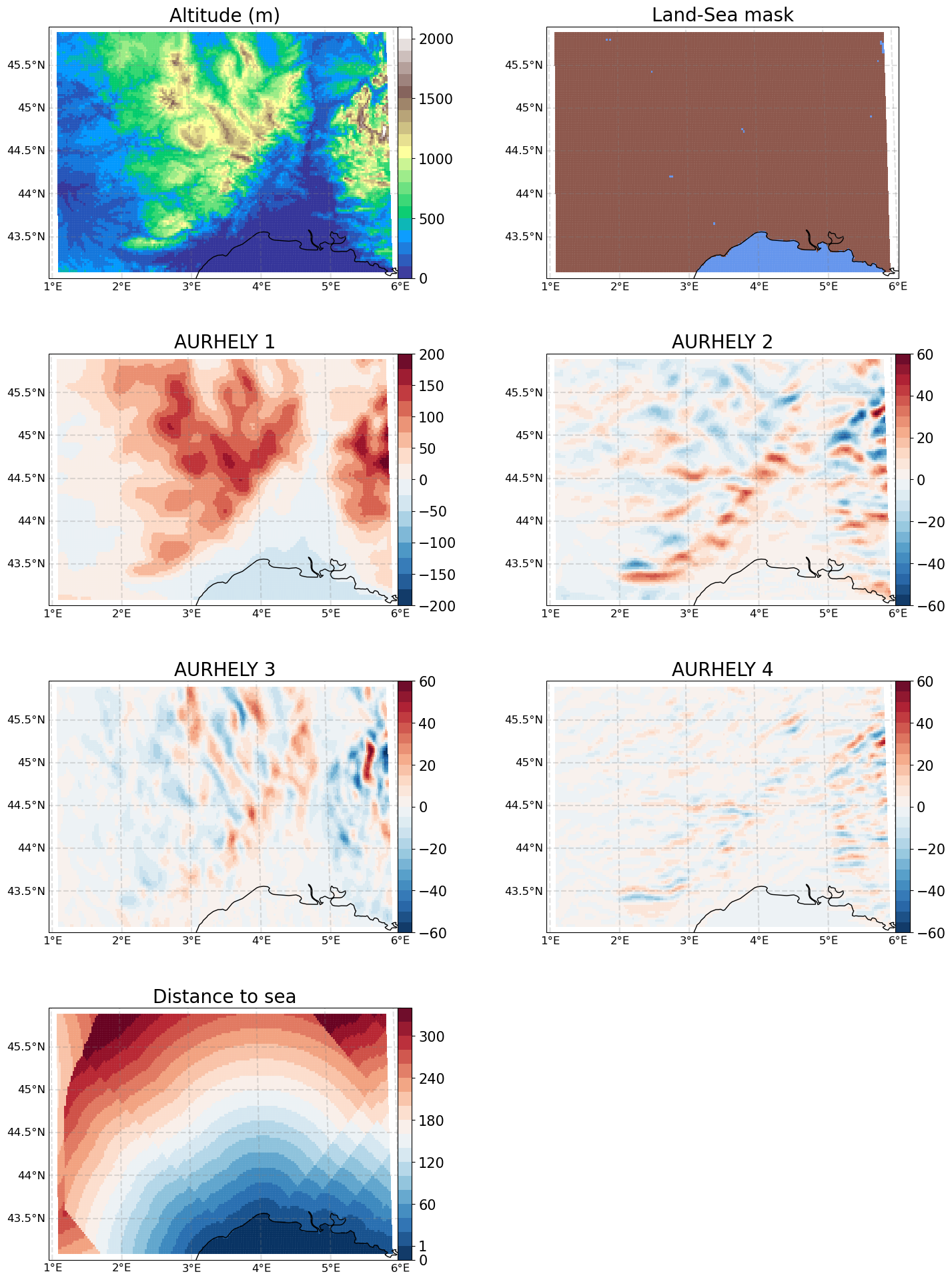}
    \caption{Constants fields used as predictors in distributional regression U-Nets: altitude, land-sea mask, four first components of AURHELY procedure, and distance to the sea.}
    \label{fig:constant-fields}
\end{figure}

In addition to summary statistics from AROME-EPS, distributional regression U-Nets (DRU) use constant fields carrying information about the topography and the type of terrain as predictors. The constant fields used are the altitude, a land-sea mask, the distance to sea and the first four components of a principal component analysis decomposition called AURHELY \citep{Benichou1994}. \cite{Lerch2022} showcased that the use of constant fields, such as altitude or orography, improves the performance of DRN. The first four components of AURHELY can be interpreted as local peak/depression, Northern/Southern slope, Eastern/Western slope and saddle effects, respectively. Figure~\ref{fig:constant-fields} shows the seven constant fields used as predictors in DRU. Table~\ref{tab:predictors} summarizes the predictors issued from both the raw ensemble and constant fields. Table~\ref{tab:dataset} lists the dimensions of the dataset. 

\def\arraystretch{1.25}%
\begin{table}[!ht]
    \centering
    \begin{tabular}{L{2cm} L{7cm}}
        \hline
        Type & \multicolumn{1}{c}{Variable}\\\hline
        \multirow{6}{*}{\rotatebox[origin=c]{90}{\parbox[c]{2.5cm}{\centering Raw ensemble\\(mean, min, max, sd)}}} & Precipitation \\
         & Convective available potential energy\\
         & Maximal reflectivity\\
         & Pseudo wet-bulb potential temperature\\
         & Relative humidity\\
         & AROME convection index\\\hline
        \multirow{4}{*}{\rotatebox[origin=c]{90}{\parbox[c]{1.8cm}{\centering Constant fields}}} & Altitude \\
        & Land-sea mask\\
        & AURHELY components (1-4)\\
        & Distance to sea\\\hline
    \end{tabular}
    \caption{List of weather and topographic variables used as predictors.}
    \label{tab:predictors}
\end{table}

\def\arraystretch{1.5}%
\begin{table}[ht]
    \centering
    \begin{tabular}{M{2cm} M{2cm} L{9cm}}
        \hline
          Variable & Value & \multicolumn{1}{c}{Description} \\\hline
          $d$ & 31 & number of predictors\\
          $H$ & 112 & height (in grid points) of the region of interest (latitude)\\
          $W$ & 192 & width (in grid points) of the region of interest (longitude)\\
          $n_{trainval}$ & 1091 & $\#$ of days in the training/validation dataset\\
          $n_{test}$ & 365 & $\#$ of days in the test dataset\\\hline
    \end{tabular}
    \caption{Dimensions of the dataset used in this study.}
    \label{tab:dataset}
\end{table}

\section{Methods}\label{section:methods}

We compare several postprocessing methods for the marginal distributions of gridded spatial ensemble forecasts of 3-h accumulated precipitation over the South of France. In a complete postprocessing scheme used operationally, the multivariate dependencies can then be retrieved using ECC or ScS, for example. We compare our U-Net-based distributional regression method to two benchmark methods: quantile regression forest (QRF; \citealt{Taillardat2016}) and QRF with tail extension (TQRF; \citealt{Taillardat2019}). The performance of postprocessed forecasts using these different methods will be compared to the performance of the raw ensemble. Additionally, we recall distributional regression networks (DRN; \citealt{Rasp2018}) since our method can be seen as an extension of this approach.

These methods differ in their degree of reliance on parametric distributions (nonparametric, semi-parametric and parametric), in the fact of being local (i.e., a different model for each grid point) or global (i.e., a single model for the whole grid). Among global methods, differences lie in the representation of the spatial structure of the data. We briefly present the benchmark techniques and their limitations.

\subsection{Quantile regression forests (QRF)}

Quantile regression forests (QRF; \citealt{Meinshausen2006}) is a nonparametric method able to predict conditional quantiles or, more generally, a conditional distribution. The method is based on random forests \citep{Breiman2001}. Similarly, it uses the data in terminal nodes (i.e., leaves) to compute a weighted average of empirical distributions. QRFs have proven their performance for postprocessing of wind speed and temperature forecasts \citep{Taillardat2016} and for precipitation forecasts \citep{Whan2018,vanStraaten2018}. QRFs can outperform complex postprocessing methods, such as neural network (NN-)based methods, at specific locations due to their local adaptability \citep{Rasp2018, Schulz2022}. Moreover, QRF is used operationally as a postprocessing method at Météo-France \citep{Taillardat2020}. This, as well as its overall performance, makes it a relevant benchmark method for this study.\\

QRFs are known to have three main limitations: potential spatial inconsistency, storage memory voracity \citep{Taillardat2020} and inability to extrapolate. The fact that QRF is a local model (i.e., a different model is used for each location, lead time, and variable) may cause problems. There is no guarantee that the output of the models is consistent spatially or temporally. Additionally, QRFs need to store the construction parameters (such as variables and thresholds of splits) of each tree of the forest and the samples used for training. This latter limitation results in the need to store a large number of parameters (especially when working with gridded data) to perform postprocessing. Lastly, QRF is incapable of extrapolating as its output is a weighted average of the training samples and does not provide a model for the distribution tail.\\

\subsection{Quantile regression forest with tail extension (TQRF)}

In order to circumvent the extrapolation inability of QRF,  semi-parametric methods based on a combination of parametric modeling and random forest were proposed. \cite{Schlosser2019} introduced distributional regression forests using maximum likelihood to infer the parameters of a censored Gaussian distribution. \cite{Taillardat2019} proposed a method using probability-weighted moments \citep{Diebolt2007} on the output of QRF to infer the parameters of an extended generalized Pareto distribution (EGPD; \citealt{Naveau2016}). The EGPD is a flexible parametric class of distributions able to jointly model the whole range of the distribution while in alignment with extreme value theory, without the requirement of threshold selection. The methods proposed in \cite{Schlosser2019}, \cite{Taillardat2019} and, more recently, \cite{Muschinski2023} can all be adapted to any suitable parametric distribution. We choose to use the semi-parametric method of \cite{Taillardat2019} based on probability-weighted moments inference.\\

Our implementation of TQRF differs from the original method described in \cite{Taillardat2019}. It uses refinements that have proven to be useful in operational settings: the tail extension is only activated if the QRF forecast assigns a large enough probability of exceedance of certain levels of interest, and in that case, only the quantiles that are higher for the fitted distribution than in the output of the QRF are updated. Moreover, we did not use EGPD because, while the QRF+EGPD is robust and efficient, the minimization of its continuous ranked probability score (CRPS; \citealt{Matheson1976}) for parameter inference is not direct due to its complex form \citep{Taillardat2019, Taillardat2019corrigendum}. These implementation issues could, for example, be circumvented by using Monte-Carlo sampling to estimate the CRPS or by fixing the tail parameter to its climatological value.

Instead of the EGPD, the generalized truncated/censored normal distribution (GTCND; \citealt{Jordan2019}) and the censored-shifted gamma distribution (CSGD; \citealt{Scheuerer2015csgd}) are used as tail extensions of the QRF and as parametric distributions for DRU. The GTCND used here has a lower endpoint equal to $0$ and no upper endpoint and its cumulative distribution function (cdf) is defined as
\begin{equation*}
    F_{L,\mu,\sigma}^\mathrm{gtcnd}(z) = \begin{cases}
        L + \frac{1-L}{1-\Phi(-\mu/\sigma)} \big(\Phi(\frac{z-\mu}{\sigma})-\Phi(-\mu/\sigma)\big) &\text{if }z\geq0\\
        0 &\text{if }z<0
    \end{cases},
\end{equation*}
where $0\leq L\leq1$ is the probability of a dry event (i.e., absence of precipitation), $\Phi$ is the cdf of the standard normal distribution, $\mu\in\bbR$ is the location parameter of the truncated normal distribution and $\sigma>0$ is its scale parameter. The cdf of the CSGD is defined as
\begin{equation*}
        F_{k,\theta,\delta}^\mathrm{csgd}(z) = \begin{cases}
        G_{k}(\frac{z-\delta}{\theta})\ &\text{if }z\geq0\\
        0\ &\text{if }z<0\\
    \end{cases},
\end{equation*}
where $G_k$ is the cdf of the gamma distribution of shape $k>0$, $\theta$ is the scale parameter and $\delta<0$ is a shift parameter. The probability of dry events has a point mass of $G_k(-\delta/\theta)$. These distributions are both suited to the forecast of precipitation since they have point masses in $0$ and take positive values. Moreover, the CSGD can reflect the variations of skewness observed in precipitation distributions \citep{Scheuerer2015csgd}. Details on the moments method for GTCND and CSGD, as well as CRPS formulas, are provided in Appendix~\ref{appendix:gtcnd} and Appendix~\ref{appendix:csgd}.\\

We denote QRF+\textit{distrib} the TQRF method where \textit{distrib} is the name of the parametric distribution family. The QRF+EGPD method is used operationally for rainfall postprocessing at Météo-France \citep{Taillardat2020}. Nonetheless, this semi-parametric method remains local and thus also suffers from both potential spatial inconsistency and memory voracity \citep{Taillardat2020}. To bypass these limitations, methods need to be global (i.e., use one model for all locations) while staying efficient locally.

\subsection{Distributional regression networks (DRN)}

\cite{Rasp2018} proposed distributional regression networks (DRN), a NN-based approach to postprocess 2-m temperature forecasts. DRN is a global model predicting the parameters of a distribution of interest. It leverages the flexibility of NN to model the dependency of parameters on the covariables (used as input of DRN). DRN can be seen as an extension of EMOS \citep{Gneiting2005}, which itself fits a parametric distribution where the parameters linearly depend on summary statistics of the raw ensemble. DRN is a global model thanks to the presence of an embedding module within its architecture, allowing the network to learn location-specific parameters and to benefit from data at similar locations. DRN learns the embedding and parameters of a dense NN by minimizing a strictly proper scoring rule \citep{Gneiting2014} such as the CRPS.

\cite{Rasp2018} and \cite{Schulz2022} have shown that DRN outperforms other state-of-the-art methods in most stations over Germany for the postprocessing of temperature and wind gusts, respectively. Moreover, \cite{Schulz2022} studied other NN-based postprocessing techniques, namely Bernstein quantile network (BQN; \citealt{Bremnes2020}) and histogram estimation network (HEN; see, e.g., \citealt{Scheuerer2020} and \citealt{Veldkamp2021}). BQN and HEN are nonparametric approaches where NNs learn the coefficient of Bernstein polynomials to predict a quantile function and probabilities of bins to predict a probability density function (pdf), respectively. At particular stations, BQN outperforms other postprocessing techniques, including DRN, for wind gust forecasts.\\

In spite of being a global model, the architecture of DRN makes it ill-suited to gridded data. Its architecture does not use knowledge of the spatial structure of the points and thus has to try to learn it through its embedding module. Moreover, DRN only uses information available at the location of interest as predictors. Convolutional neural network (CNN)-based architectures make use of the gridded structure of the data and can use the information at neighboring locations as a predictor. \cite{Lerch2022} studied a modified DRN architecture using the representation of global fields from a convolutional auto-encoder as predictors and showed an improvement in skill compared to regular DRN.\\

DRNs' architecture makes their implementation on gridded data very costly. They need to flatten the data across locations (i.e., reshape it into a 1D vector), and they cannot benefit from GPU computing. For these reasons and their impact on the search for optimal hyperparameters, DRNs are not used as a benchmark method in this study.

\subsection{Distributional regression U-Nets (DRU)}

Convolutional blocks are the main ingredient of CNN-based architectures. The simplest convolutional blocks are composed of a convolutional layer and a max-pooling layer. The role of the convolutional layer is to learn kernels able to extract useful features from the input of the convolutional block. The max-pooling layer reduces the resolution of the features, allowing the following layers to work at broader scales. The succession of convolutional blocks allows CNNs to learn patterns at different spatial scales and to learn complex patterns (see, e.g., \citealt{Simonyan2014}). CNN-based architectures have been used in numerous postprocessing studies (e.g., \citealt{Dai2021, Veldkamp2021, Li2022, Chapman2022, Lerch2022}).\\

Since we are interested in global models using the data's gridded structure and want the output to be the distributional parameters of marginals on the same grid, we use a U-Net architecture \citep{Ronneberger2015}. The U-Net architecture was initially designed for images but is compatible with gridded data to obtain a grid-based output. It has been used for various postprocessing applications. \cite{Groenquist2021} used it in a bias/uncertainty postprocessing scheme of temperature and geopotential forecasts. \cite{Dai2021} used a U-Net as a generator within a conditional generative adversarial network (cGAN) for cloud cover postprocessing. \cite{Hu2023} used U-Nets to predict the parameters of a CSGD corresponding to the postprocessed daily precipitation given a deterministic forecast. \cite{Horat2023} used U-Nets to perform postprocessing of temperature and precipitation at the sub-seasonal to seasonal scale. The task is a three-level classification problem with below-normal, near-normal and above-normal conditions as classes. \cite{BenBouallegue2024} used transformers within a U-Net architecture to postprocess ensemble members directly with temperature and precipitation as variables of interest.\\

\begin{figure}[ht]
    \centering
    \includegraphics[width=\textwidth]{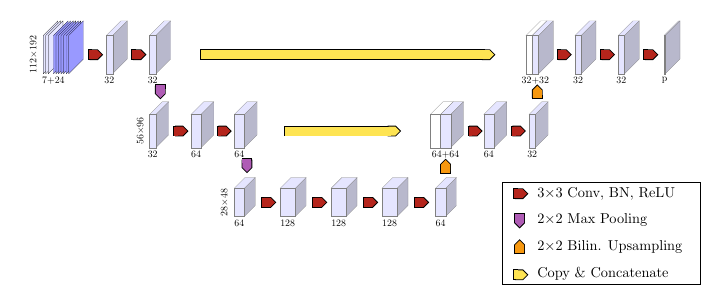}
    \caption{Architecture of distributional regression U-Nets. \textit{Conv} stands for convolution, \textit{BN} stands for batch normalization, \textit{ReLU} stands for rectified linear unit and \textit{Bilin. Upsampling} stands for bilinear upsampling. $p$ is the number of distribution parameters: for GTCND and CSGD, $p=3$.}
    \label{fig:u-net}
\end{figure}

The U-Net architecture used in this work is presented in Figure~\ref{fig:u-net}. The U-Net input is a concatenation of constant fields and summary statistics of the ensemble members. The output is the parameters of the postprocessed marginal distribution at each grid point (i.e., parameters of a GTCND or a CSGD). The architecture can be decomposed into two parts. On the left part, the succession of specific convolutional blocks (red and purple arrows) leads to an increase in the number of features and a reduction of the spatial dimension (i.e., a coarsening of the spatial resolution) as the data progresses through the network. As explained above, the convolutional blocks are constructed in order to learn useful representations of the features of the fields at various spatial scales. On the right part, upscaling blocks (red and orange arrows), based on bilinear upsampling, use the features learned in the central part of the architecture to predict features at finer resolutions and finally learn the parameters of the distribution selected. Additionally, we use skip-connections (yellow arrows), consisting of copying and concatenating features, as bridges between the left and right parts of the U-Net. Skip-connections have proven to improve the stability of the convergence of NN (see, e.g., \citealt{Li2018}). This U-Net-based method is a global model enabling extrapolation through a parametric distribution (e.g., GTCND or CSGD). We denote U-Net+\textit{distrib} the distributional regression U-Net (DRU) where \textit{distrib} is the parametric distribution. \\

DRU learns to predict the parameters of a distribution by minimizing the CRPS at each grid point. Both the parameterized distribution and the scoring rule to minimize can be chosen to be suited to the variable of interest or to facilitate computations, thus making the architecture flexible. The convolution blocks allow the parameters of marginal distribution to be learned from neighboring grid points, potentially accounting for dependencies between grid points \citep[Section~4.5]{Schefzik2018}. Moreover, the use of constant fields as input enables the convolutional layers to learn representations of these fields that are relevant to the postprocessing task at hand. This can be seen as a natural extension of the embedding module in DRN \citep{Rasp2018}.\\

DRNs are built to bypass the limitations of the methods presented above. The model is global and uses the predictor fields of the whole grid, this construction enables the predicted marginals to be spatially consistent. Moreover, the use of convolutional layers facilitates the learning of relevant spatial features compared to DRN. Memory voracity is not an issue as the model is global and the number of parameters is contained. Finally, as highlighted previously, any parameterized distribution can be used as the output of DRU accounting for extrapolation and relevance to the target variable at hand. Table~\ref{tab:benchmark} summarizes the characteristics of the postprocessing methods studied in this article.\\

The U-Net-based method of this article is related to the one of \cite{Hu2023} in the sense that both approaches use U-Nets to predict the parameters of a distribution corresponding to the marginals of the variable of interest. The main differences between the approaches are the following: they studied daily precipitation accumulations, where we are interested in 3-h accumulated precipitation; they postprocess deterministic forecasts, where we postprocess ensemble forecasts; and finally, we use constant fields as additional predictors. Moreover, in terms of the number of years in the training data, our work (with only 3 years of training data) falls in a "gray area" where their U-Net-based method is outperformed by analog ensemble \citep{DelleMonache2013}, which is a simpler approach \citep[Figure 11]{Hu2023}. Table~\ref{tab:unets} summarizes the characteristics of the different U-Net-based postprocessing methods available.\\

\def\arraystretch{1.5}%
\begin{table}[ht]
    \centering
    \begin{tabular}{|M{2cm}|M{2cm}|M{2cm}|M{3cm}|M{3cm}|}
        \cline{2-5}
        \multicolumn{1}{c|}{} & QRF & TQRF & DRN & DRU \\\hline
        Local/Global & local & local & global & global \\\hline
        Principles & grid point per grid point & grid point per grid point & embedding to learn from similar stations & constant fields and architecture aware of the gridded structure\\\hline
        Ability to extrapolate & \xmark & \cmark & \cmark & \cmark \\\hline
        Number of parameters & $\sim$15.3~B & $\sim$15.3~B & $\sim$450,000 & $\sim$1,000,000 \\\hline
        Storage necessary for prediction & splits of each tree and training data & splits of each tree and training data & parameters and architecture & parameters and architecture\\\hline
    \end{tabular}
    \caption{Comparison of the postprocessing methods mentioned in this study. The number of parameters is provided for hyperparameters selected by cross-validation on the training/validation data set and for the setup described in Section~\ref{section:data} (e.g., a $112\times192$ grid). In the case of DRN, an architecture similar to the one in \cite{Rasp2018} has been considered. \textit{B} stands for billion.}
    \label{tab:benchmark}
\end{table}

\newcommand{\pad}[1]{\vspace{.75em}{\small #1}\vspace{.75em}}
\def\arraystretch{1.25}%
\begin{table}[ht]
    \begin{adjustwidth}{-1cm}{-1cm}
        \small
        \centering
        \begin{tabular}{|M{.75em} M{2em}|M{6em}|M{6em}|M{6em}|M{6.5em}|M{6em}|M{5.5em}|}
        \cline{3-8}
        \multicolumn{2}{c|}{} & \cite{Groenquist2021} & \cite{Dai2021} & \cite{Horat2023} & \cite{BenBouallegue2024} & \cite{Hu2023} & Pic et al. (2024) \\\hline
        \multicolumn{2}{|M{4.5em}|}{Variable of interest} & temperature, geopotential & cloud cover & temperature, 2-w precip. & temperature, 6-h precip. & 24-h precip. & 3-h precip. \\\hline
        \multicolumn{2}{|M{4.5em}|}{Output} & bias & samples from a cGAN & probability of classes & postprocessed ensemble members & parameters of a CSGD & parameters of a \textsc{gtcnd/csgd} \\\hline
        \multicolumn{2}{|M{4.5em}|}{Lead times} & 48h & 1-120h & 2-4w & 6-96h & 0-4d & 21h \\\hline
        \multirow{2}{*}{\rotatebox[origin=c]{90}{\parbox[c]{1.8cm}{\centering Dataset}}} & \rotatebox[origin=c]{90}{\parbox[c]{1.25cm}{\centering raw forecast}}\vspace{.25em} & \pad{ECMWF-ENS10 \textit{ensemble}} & \pad{COSMO-E, ECMWF-IFS \textit{ensemble}} & \pad{ECMWF-IFS (S2S) \textit{ensemble}} & \pad{ECMWF-IFS  \textit{ensemble}} & \pad{West-WRF \textit{deterministic}} & \pad{AROME-EPS \textit{ensemble}}\\\cline{2-2}
        & \vspace{.25em} \rotatebox[origin=c]{90}{\parbox[c]{.5cm}{\centering obs.}}\vspace{.25em} & \pad{ERA5} & \pad{EUMETSAT} & \pad{NOAA-CPC} & \pad{ERA5} & \pad{PRISM} & \pad{ANTILOPE} \\\hline
        \multicolumn{2}{|M{4.5em}|}{Resolution} & 0.5° & 0.02° & 1.5° & 1° & 0.04° & 0.025° \\\hline
        \multicolumn{2}{|M{4.5em}|}{Training data range} & 17~years & 3~years & 20~years & 19~years & 2-30~years & 3~years \\\hline
        \end{tabular}
    \end{adjustwidth}
    \caption{Comparison of the postprocessing methods relying on U-Nets.}
    \label{tab:unets}
\end{table}

The following hyperparameters of the U-Net architecture have been selected using the training/validation dataset: the learning rate, the batch size and the number of epochs. The optimizer is Adam with default parameters (except for the learning rate) from its \texttt{Keras} implementation. In order to limit the number of parameters and prevent overfitting, the depth of the U-Net is kept at two levels (as shown in Fig.~\ref{fig:u-net}) and separable convolutions were used instead of standard ones. Moreover, in order to contain the variability due to random initialization, we aggregate forecast distributions of 10 models as recommended in \cite{Schulz2022aggregating}. 

Most of the implementation was conducted in \texttt{Python} and the implementation of DRU is based on \texttt{Tensorflow} \citep{Tensorflow2015} and \texttt{Keras} \citep{Chollet2015}. QRF and TQRF are implemented in \texttt{R} \citep{RCT2023} using the \texttt{ranger} package \citep{Wright2017}.

\section{Results}\label{section:results}

We provide a comparison of DRU to QRF, TQRF and the raw ensemble using verification tools targeting three different aspects of forecasts: verification of the overall performance with the CRPS, calibration and extreme events. First, we compare the performance of the postprocessing techniques in terms of their relative improvement compared to the raw ensemble and among themselves. This improvement is quantified in terms of continuous ranked probability skill score (CRPSS). Second, we assess the calibration of the postprocessed forecasts using rank histograms. Finally, the improvement of the postprocessing methods in terms of extreme forecasting is evaluated using receiver operating characteristic (ROC) curves for events corresponding to the exceedance of various thresholds.

\subsection{Continuous ranked probability score}

Since the postprocessing techniques considered act on the 1-dimensional marginals, the improvement and comparison of the postprocessing techniques can be done with univariate scoring rules. The continuous ranked probability score (CRPS; \citealt{Matheson1976}) is one of the most popular univariate scoring rules in weather forecasting and is defined as
\begin{align}
    \CRPS(F,y) &= \int_\bbR (F(z)-\mathds{1}_{y\leq z})^2 \rmd z;\label{eq:CRPS_bs}\\ 
               &= 2 \int_0^1 (\mathds{1}_{y\leq F^{-1}(\alpha)}-\alpha)(F^{-1}(\alpha)-y) \rmd\alpha;\label{eq:CRPS_qs}\\
               &= \bbE_F|X-y|-\frac{1}{2}\bbE_F|X-X'|,\label{eq:CRPS_kernel}
\end{align}
where the forecast $F$ is assimilated to its cdf, $F^{-1}$ is its quantile function and $X$ and $X'$ follow the distribution $F$. The CRPS is strictly proper on the set of measures with a finite first moment. Moreover, it benefits from multiple representations that help both its computation and interpretation. Equation~\eqref{eq:CRPS_bs} is the threshold or Brier score \citep{Brier1950} representation and expresses the CRPS as the integrated squared error between the cdf of the forecast and the empirical cdf associated with observation $y$ over all thresholds $z$. Equation~\eqref{eq:CRPS_qs} is the quantile representation and shows that the CRPS is expressed as the pinball loss over all quantile levels $\alpha$. Equation~\eqref{eq:CRPS_kernel} is the kernel representation and is particularly useful to compute the score of ensemble forecasts. The CRPS formulas for the parametric distributions of this article are available in the Appendix~\ref{appendix:gtcnd} and \ref{appendix:csgd}. For the raw ensemble, QRF and TQRF forecasts, the CRPS has been estimated using the fair estimator \citep{Ferro2013}.\\

When working with (strictly) proper scoring rules to compare forecasts, the comparison of the scoring rules of two forecasts can be summarized by the skill score. For a proper scoring rule $\rmS$, the skill score of a forecast $F$ with respect to (w.r.t.) a reference forecast $F_\mathrm{ref}$ is defined as
\begin{equation}
    \mathrm{SS}(F,F_\mathrm{ref}) = \cfrac{\bbE_G[\rmS(F_\mathrm{ref},Y)]-\bbE_G[\rmS(F,Y)]}{\bbE_G[\rmS(F_\mathrm{ref},Y)]},
\end{equation}
where $G$ is the distribution of the observations and $\bbE_G[\cdots]$ is the expectation with respect to $Y\sim G$. The skill score is positive if the forecast $F$ improves the expected score w.r.t. the reference forecast $F_\mathrm{ref}$ and negative otherwise. The skill score can be expressed in percentage. In the context of postprocessing, a reference of choice is the raw ensemble that the postprocessing procedure aims to improve upon. \\

\def\scalecrps{.275}
\begin{figure}[ht]
    \centering
    \begin{subfigure}[c]{.5\textwidth}
        \centering
        \includegraphics[scale=\scalecrps]{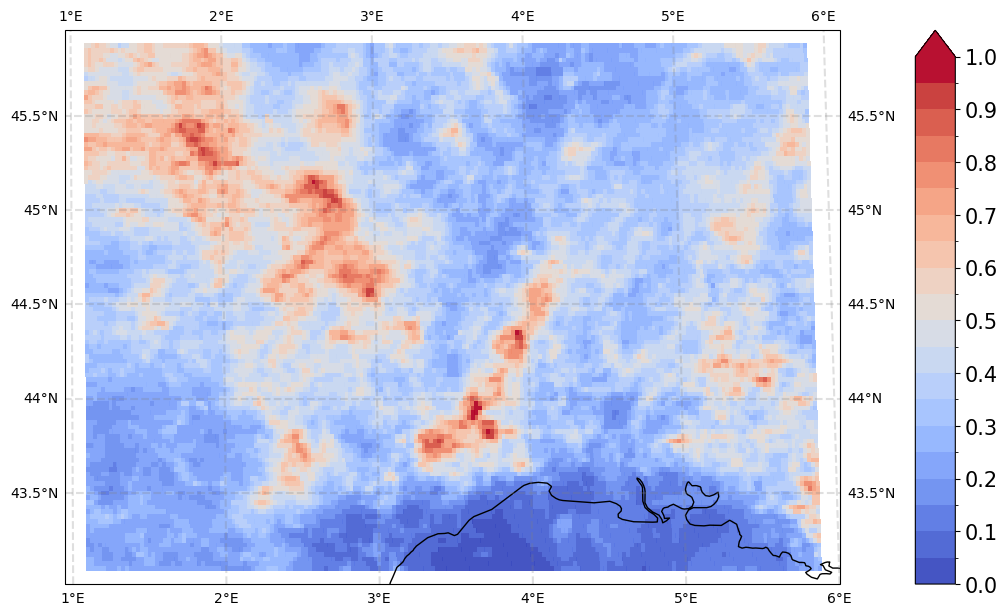}
        \caption{Expected CRPS of raw ensemble}\label{fig:crps_raw}
    \end{subfigure}%
    \begin{subfigure}[c]{.5\textwidth}
        \centering
        \includegraphics[scale=\scalecrps]{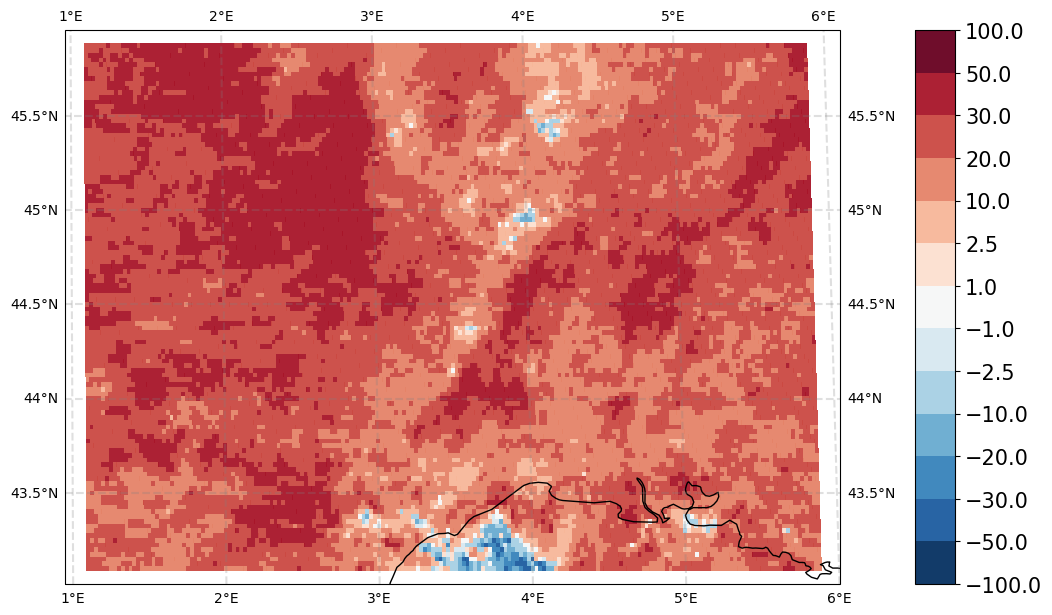}
        \caption{CRPSS of QRF w.r.t. raw}\label{fig:crpss_qrf_raw}
    \end{subfigure}
    
    \begin{subfigure}[c]{.5\textwidth}
        \centering
        \includegraphics[scale=\scalecrps]{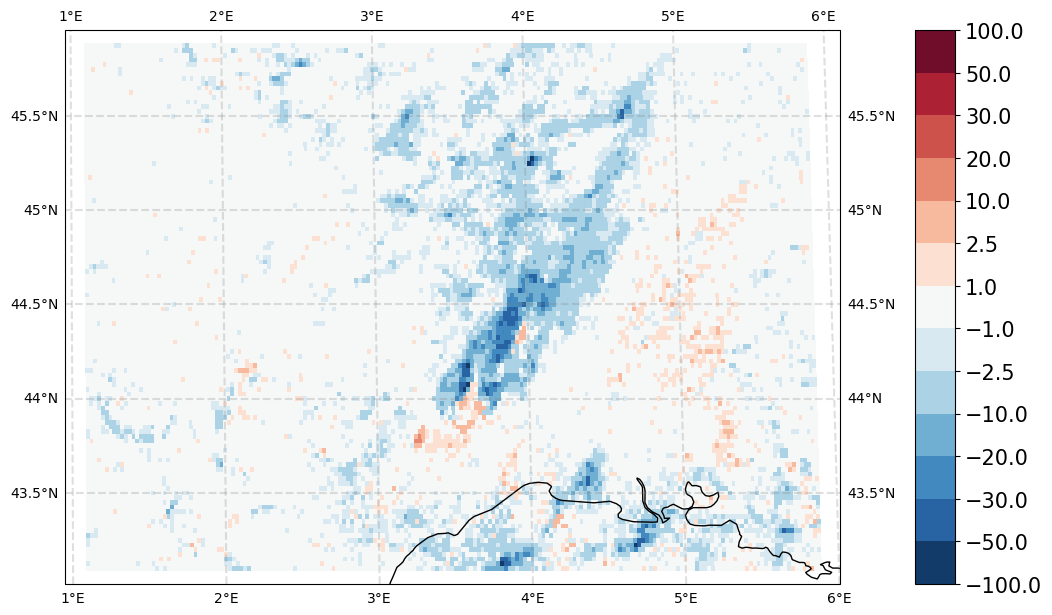}
        \caption{CRPSS of QRF+GTCND w.r.t. QRF}\label{fig:crpss_qrf+gtcnd_qrf}
    \end{subfigure}%
    \begin{subfigure}[c]{.5\textwidth}
        \centering
        \includegraphics[scale=\scalecrps]{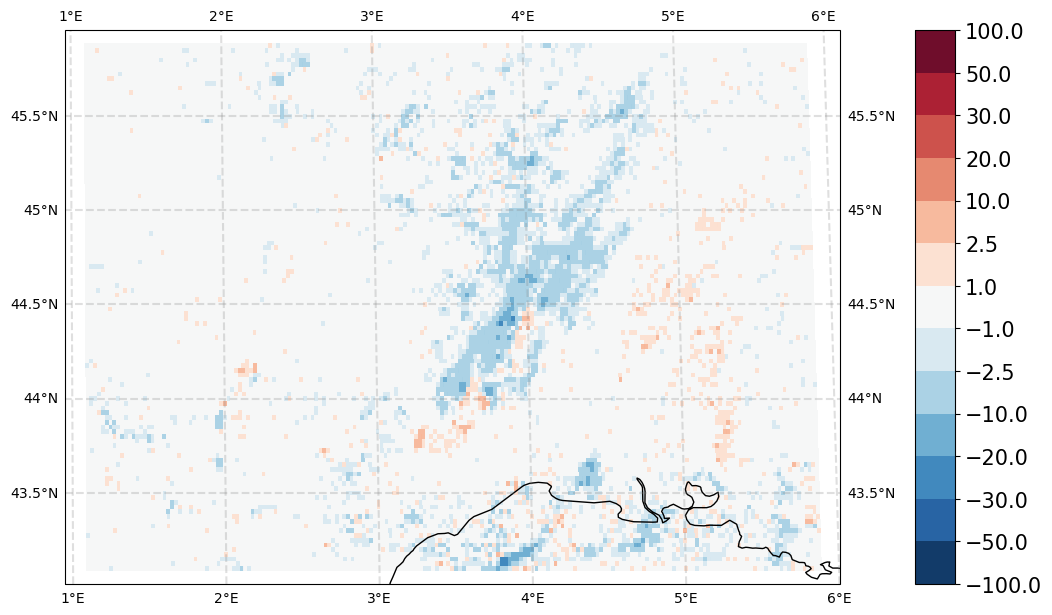}
        \caption{CRPSS of QRF+CSGD w.r.t. QRF}\label{fig:crpss_qrf+csgd_qrf}
    \end{subfigure}
    \caption{Predictive performance of the benchmark methods in terms of CRPS. (a) Expected CRPS of the raw ensemble, (b) CRPSS of QRF w.r.t. the raw ensemble and CRPSS w.r.t. QRF of (c) QRF+GTCND and (d) QRF+CSGD. }
    \label{fig:crps_benchmark}
\end{figure}

\def\scalecrps{.275}
\begin{figure}[!ht]
    \centering
    \begin{subfigure}[c]{.5\textwidth}
        \centering
        \includegraphics[scale=\scalecrps]{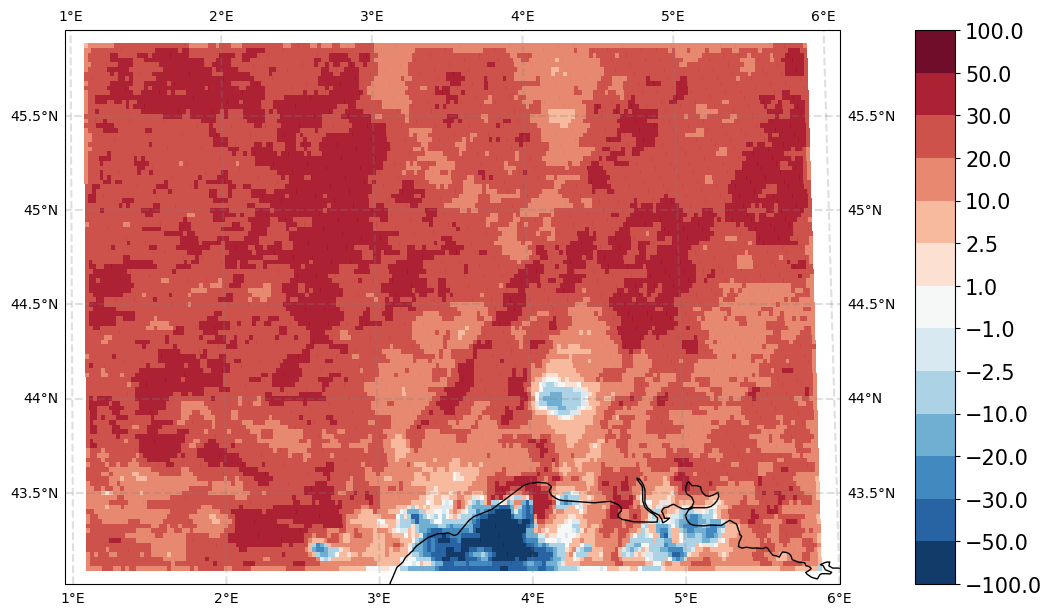}
        \caption{CRPSS of U-Net+GTCND w.r.t. raw}\label{fig:crpss_unet+gtcnd_raw}
    \end{subfigure}%
    \begin{subfigure}[c]{.5\textwidth}
        \centering
        \includegraphics[scale=\scalecrps]{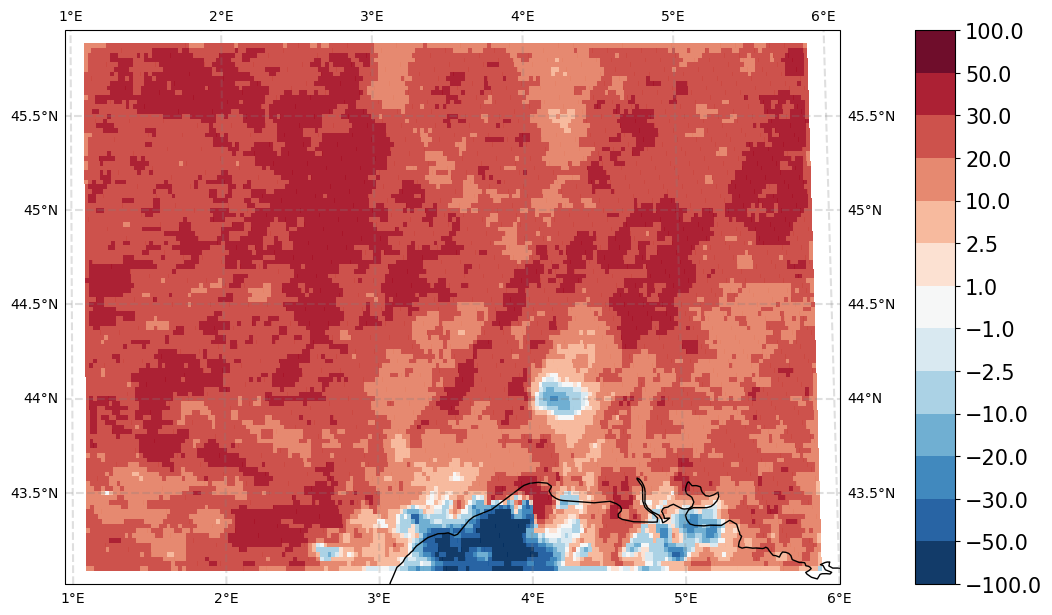}
        \caption{CRPSS of U-Net+CSGD w.r.t. raw}\label{fig:crpss_unet+csgd_raw}
    \end{subfigure}
    
    \begin{subfigure}[c]{.5\textwidth}
        \centering
        \includegraphics[scale=\scalecrps]{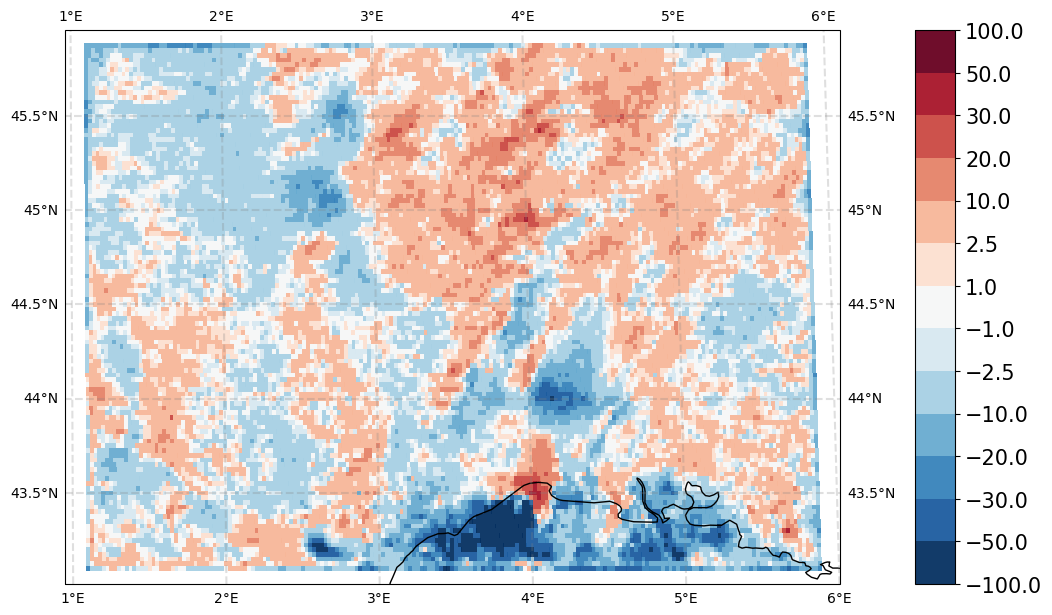}
        \caption{CRPSS of U-Net+GTCND w.r.t. QRF}\label{fig:crpss_unet+gtcnd_qrf}
    \end{subfigure}%
    \begin{subfigure}[c]{.5\textwidth}
        \centering
        \includegraphics[scale=\scalecrps]{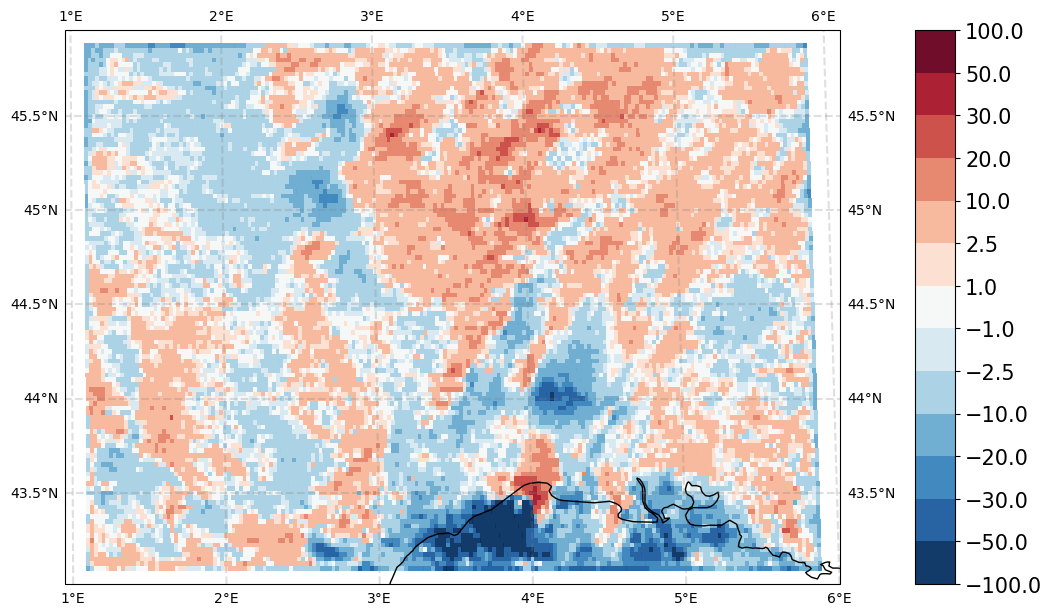}
        \caption{CRPSS of U-Net+CSGD w.r.t. QRF}\label{fig:crpss_unet+csgd_qrf}
    \end{subfigure}
    \caption{Predictive performance of the distributional regression U-Nets in terms of CRPS. CRPSS w.r.t. the raw ensemble of (a) U-Net+GTCND and (b) U-Net+CSGD and CRPSS w.r.t. QRF of (c) U-Net+GTCND and (d) U-Net+CSGD.}
    \label{fig:crps_unet}
\end{figure}

\begin{figure}[!h]
    \centering
    \includegraphics[width=.95\textwidth]{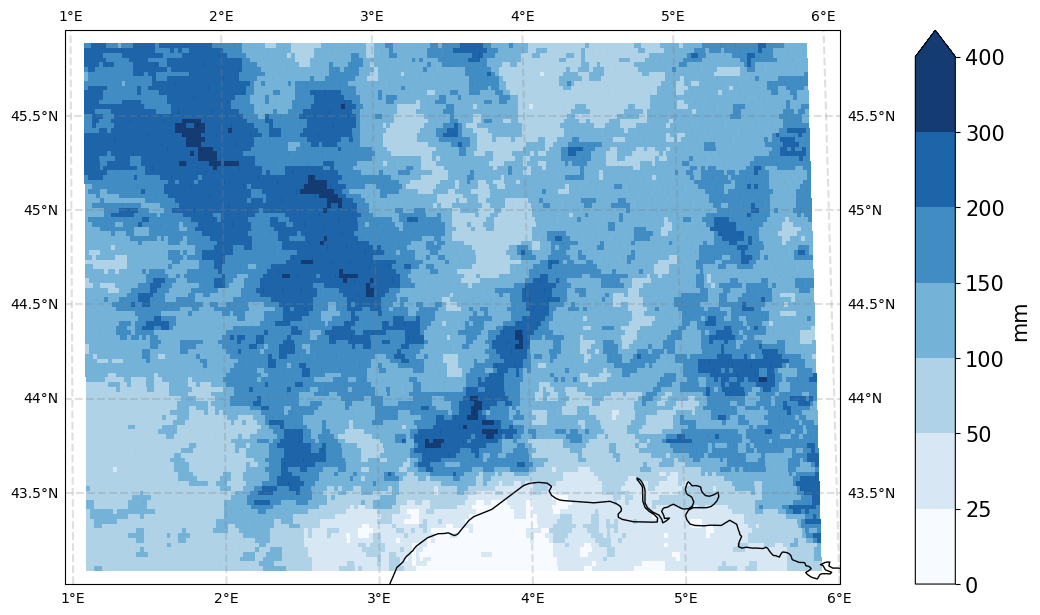}
    \caption{Total precipitation over the test set. Due to the initial time and the lead time considered, only precipitation between 12:00UTC and 15:00UTC are taken into account.}
    \label{fig:total_precip_test}
\end{figure}

We compared the continuous ranked probability skill score (CRPSS) for the different postprocessing methods studied w.r.t. other benchmark methods. Figure~\ref{fig:crps_benchmark} shows the expected CRPS of the raw ensemble, the CRPSS of QRF w.r.t. the raw ensemble and the CRPSS of QRF+GTCND and QRF+CSGD w.r.t. QRF. The raw ensemble has an expected CRPS of $0.3725$~mm when averaged over the whole region of interest. However, the expected CRPS greatly fluctuates over the whole grid and most grid points of higher altitude have larger expected CRPS since they correspond to higher precipitation accumulations (see Fig.~\ref{fig:crps_raw}). The lowest expected CRPS values are located over the Mediterranean Sea corresponding to an area of low precipitation as discussed further (see Fig.~\ref{fig:total_precip_test}). Moreover, observations in this area are of lower quality since it is far from the nearest radar and cannot be corrected by gauges.

Figure~\ref{fig:crpss_qrf_raw} confirms that QRF is able to improve the predictive performance in terms of CRPSS compared to the raw ensemble ($23.51$\% after averaging over the region of interest). The CRPSS of QRF w.r.t. the raw ensemble is positive (i.e., improvement of skill) over the whole domain except for some localized regions. In particular, over the area that has the lowest expected CRPS for the raw ensemble, QRF is not able to improve compared to raw ensemble in terms of expected CRPS. This may be caused by the fact that this area is already well-predicted by the raw ensemble and the QRF is not able to improve its CRPS. Figures~\ref{fig:crpss_qrf+gtcnd_qrf} and \ref{fig:crpss_qrf+csgd_qrf} show the CRPSS w.r.t. QRF of QRF+GTCND and QRF+CSGD, respectively. Overall, QRF+GTCND and QRF+CSGD have a close but slightly smaller expected CRPS than that of QRF (average CRPSS w.r.t. QRF of $-1.04$\% and $-0.33$\%, respectively). For both GTCND and CSGD tail extensions, the areas of lower skill (in blue) are located in a mountainous region (the Eastern part of Massif Central) and near the Mediterranean coast. Nonetheless, the areas are wider and have lower CRPSS values for QRF+GTCND compared to QRF+CSGD. Both methods also present areas of improvement of CRPSS (in orange/red) that are sparser and smaller than the areas of negative CRPSS.\\

\def\arraystretch{1.5}%
\begin{table}[ht]
    \centering
    \begin{tabular}{|M{1cm} M{2.75cm}|M{2cm}|M{2cm}||M{2cm}|M{2cm}|}
    \cline{3-6}
        \multicolumn{2}{c|}{} & \multicolumn{4}{c|}{Reference} \\
        \multicolumn{2}{c|}{} & \multicolumn{2}{c}{Full region} & \multicolumn{2}{c|}{Censored region} \\
         \multicolumn{2}{c|}{} & Raw ensemble & QRF & Raw ensemble & QRF \\\hline
          \multirow{5}{*}{\rotatebox[origin=c]{90}{\parbox[c]{2.5cm}{\centering Postprocessing methods}}} & QRF & \textbf{23.51\%} & -- & 23.56\% & -- \\\cline{2-2}
          & QRF+GTCND & 22.67\% & -1.04\% & 22.72\% & -1.05\% \\\cline{2-2}
          & QRF+CSGD & 23.23\% & -0.33\% & 23.29\% & -0.34\% \\\cline{2-2}
          & U-Net+GTCND & 22.25\% & -1.52\% & 24.34\% & 0.05\% \\\cline{2-2}
          & U-Net+CSGD & 22.36\% & -1.37\% & \textbf{24.48\%} & \textbf{0.26\%} \\\hline
    \end{tabular}
    \caption{Summary of the performance in terms of CRPSS averaged over the full region of interest and over the censored one.}
    \label{tab:CRPSS}
\end{table}

\def\scaleforecast{.295}
\begin{figure}[ht]
    \centering
    \begin{subfigure}[c]{.5\textwidth}
        \centering
        \includegraphics[scale=\scaleforecast]{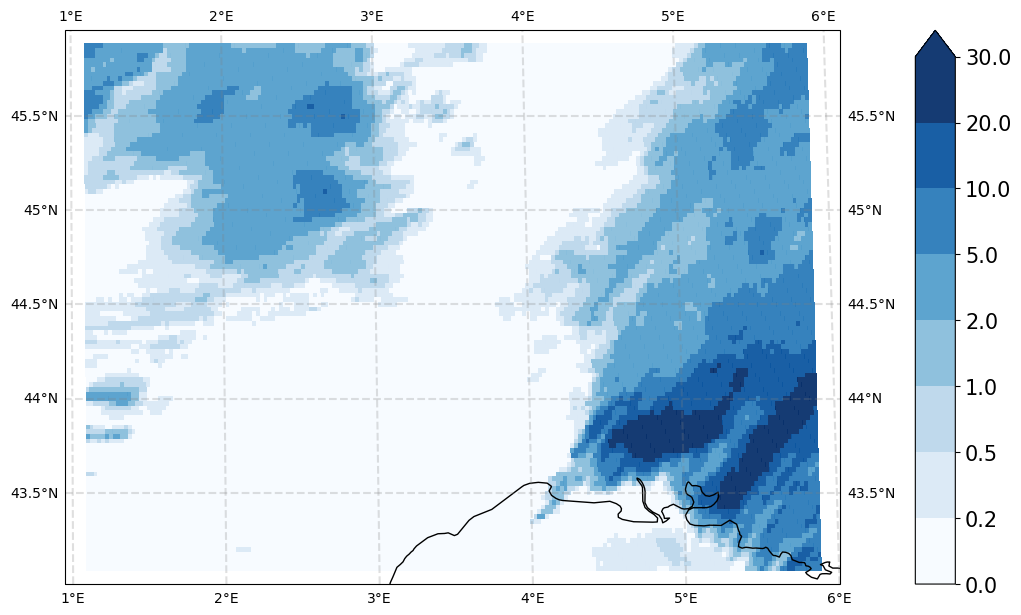}
        \caption{Precipitation (mm)}\label{fig:example_forecast_obs}
    \end{subfigure}%
    \begin{subfigure}[c]{.5\textwidth}
        \centering
        \includegraphics[scale=\scaleforecast]{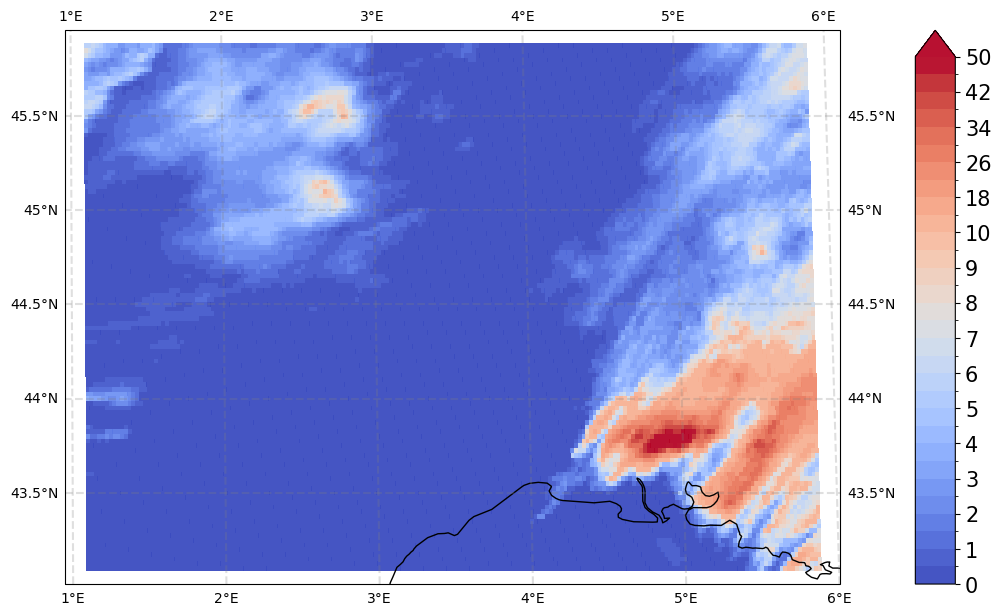}
        \caption{CRPS of raw ensemble}\label{fig:example_forecast_CRPS_raw}
    \end{subfigure}
    
    \begin{subfigure}[c]{.5\textwidth}
        \centering
        \includegraphics[scale=\scaleforecast]{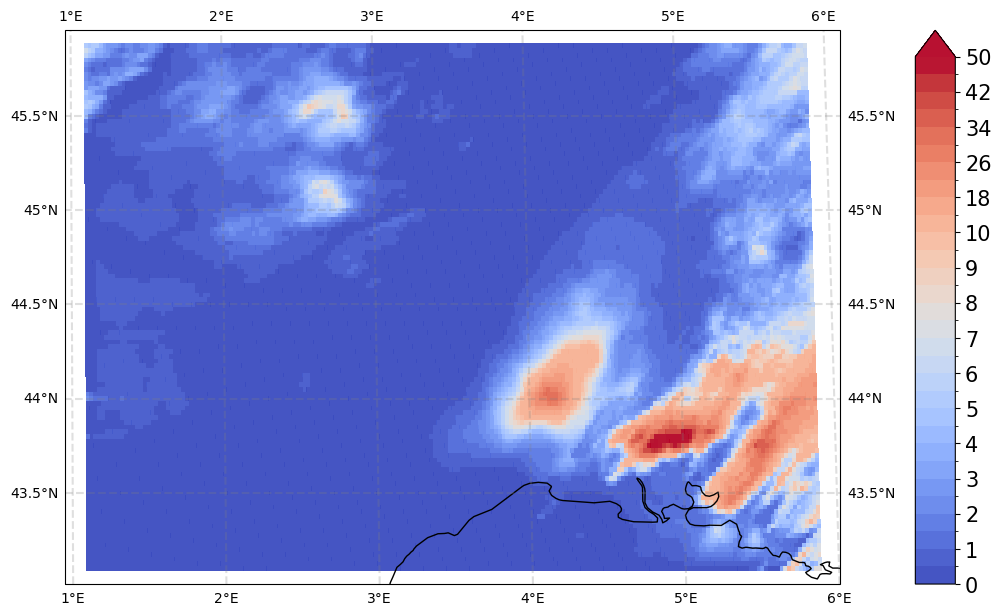}
        \caption{CRPS of U-Net+GTCND}\label{fig:example_forecast_CRPS_unet+gtcnd}
    \end{subfigure}%
    \begin{subfigure}[c]{.5\textwidth}
        \centering
        \includegraphics[scale=\scaleforecast]{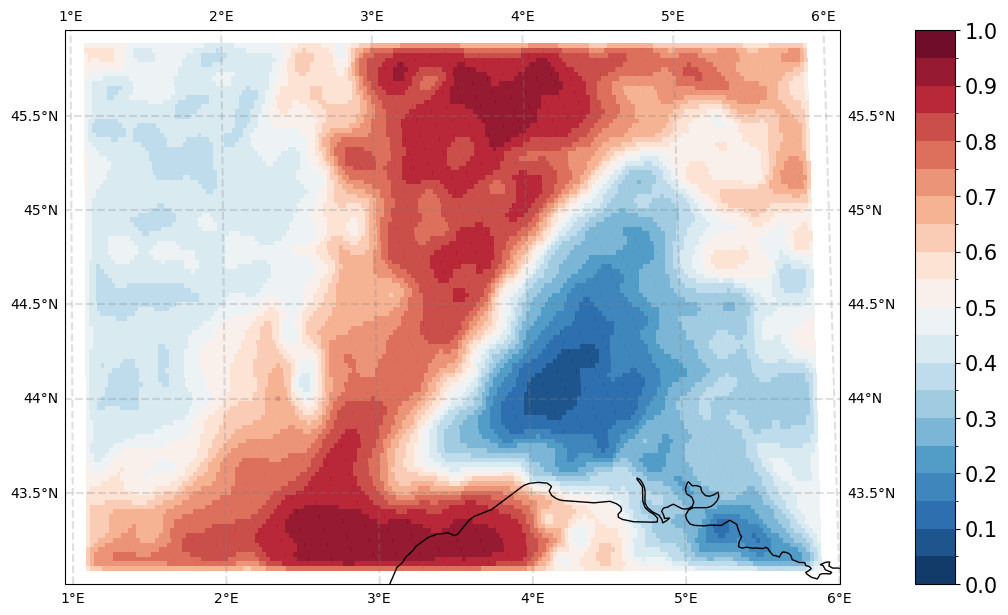}
        \caption{Predicted $L$}\label{fig:example_forecast_L}
    \end{subfigure}
    
    \begin{subfigure}[c]{.5\textwidth}
        \centering
        \includegraphics[scale=\scaleforecast]{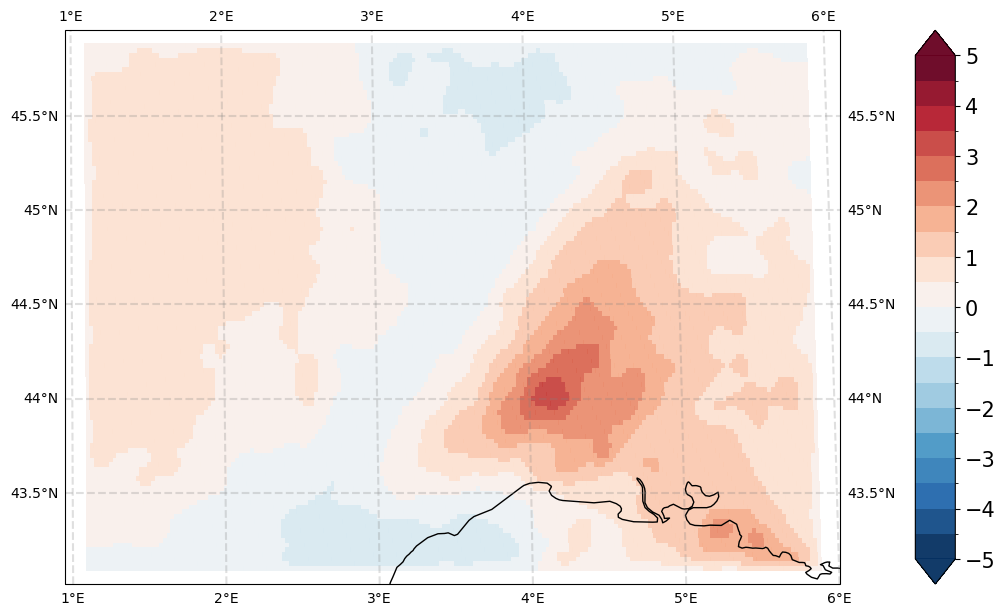}
        \caption{Predicted $\mu$}\label{fig:example_forecast_mu}
    \end{subfigure}%
    \begin{subfigure}[c]{.5\textwidth}
        \centering
        \includegraphics[scale=\scaleforecast]{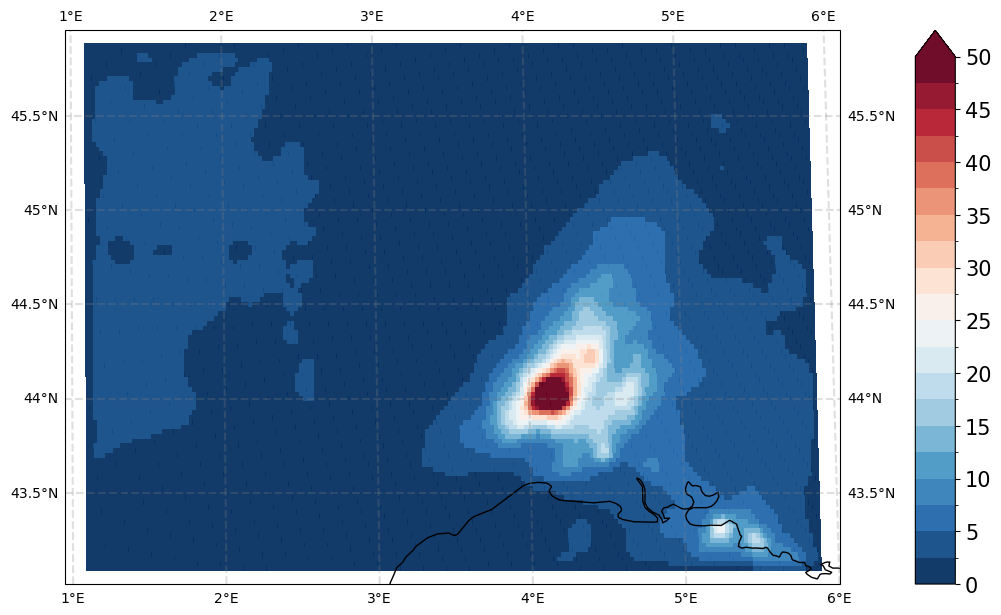}
        \caption{Predicted $\sigma$}\label{fig:example_forecast_sigma}
    \end{subfigure}

    \caption{Example of a numerical instability of U-Net+GTCND for a forecast valid on November 3, 2022 at 15
    :00UTC. Note the different scales for CRPS below and above 10~mm.}
    \label{fig:example_forecast_error}
\end{figure}

Figure~\ref{fig:crps_unet} provides the CRPSS of U-Net+GTCND and U-Net+CSGD w.r.t. the raw ensemble and QRF. Figures~\ref{fig:crpss_unet+gtcnd_raw} and \ref{fig:crpss_unet+csgd_raw} show the CRPSS of DRU w.r.t. the raw ensemble. Both GTCND and CSGD lead to methods improving CRPSS w.r.t. the raw ensemble with $22.28$\% and $22.36$\%, respectively, when averaged over the region of interest. As the QRF, DRU leads to improvement in terms of CRPSS over the vast majority of grid points. Nonetheless, there are areas where they have a poorer predictive performance compared to raw ensemble. These areas are also located over the Mediterranean Sea or near the coast, and one patch is located in the Rhône River valley. When censoring grid points located over the sea and at the border, the average CRPSS w.r.t. the raw ensemble is $24.34$\% and $24.48$\% for U-Net+GTCND and U-Net+CSGD, respectively. 

Figures~\ref{fig:crpss_unet+gtcnd_qrf} and \ref{fig:crpss_unet+csgd_qrf} show the CRPSS of U-Net+GTCND and U-Net+CSGD w.r.t. QRF. Overall, DRU has a higher expected CRPS than QRF (CRPSS of $-1.52$\% for the U-Net+GTCND and $-1.37$\% for the U-Net+CSGD), but it has an improved predictive performance (in terms of CRPS) over a non-negligible part of the region of interest. Due to their architecture, DRUs are affected by a border effect, leading to a less predictive performance on the grid points located at the boundaries of the grid (see Fig.~\ref{fig:crpss_unet+gtcnd_qrf} and Fig.~\ref{fig:crpss_unet+csgd_qrf}). Using the censoring mentioned above, U-Net+GTCND and U-Net+CSGD have an average CRPSS w.r.t. QRF of $0.05$\% and $0.26$\%, respectively. Table~\ref{tab:CRPSS} summarizes the comparisons of methods in terms of CRPSS.\\

For the training/validation dataset, DRUs are prone to numerical instabilities. This led to areas of negative CRPSS w.r.t. the raw ensemble caused by the divergence of predicted parameters ($\sigma$ is the case of U-Net+GTCND and $\theta$ in the case of U-Net+CSGD) (not shown). In addition to standard numerical stabilizing tricks, we have tried to constrain the range of diverging parameters using the value of the climatological fits since higher values would lead to forecasts less informative than the climatological forecasts. This solved the divergence issues over both the training/validation and test datasets for U-Net+CSGD but not for U-Net+GTCND (not shown). However, it increased the border effects causing deteriorating performance for both models. Hence, the constraining of the range of the parameters for DRU method is not used and the numerical stability of the methods needs to be understood and prevented.

Despite being prone to numerical instabilities, the areas of negative CRPSS w.r.t. the raw ensemble for the test dataset are not all caused by numerical instabilities. The largest area of negative CRPSS w.r.t. raw (see Fig.~\ref{fig:crpss_unet+gtcnd_raw} and \ref{fig:crpss_unet+csgd_raw}) coincide with the area with the lowest total precipitation over the test period (see Fig.~\ref{fig:total_precip_test}). This area matches the area of the lowest expected CRPS for the raw ensemble (see Fig.~\ref{fig:crps_raw}). Numerous dry events occur at this location and are perfectly predicted by the raw ensemble (i.e., all members predict 0~mm of precipitation). However, in order to perfectly predict a dry event, U-Net+GTCND and U-Net+CSGD need to predict $L=1$ and $-\delta/\theta=\infty$, respectively, which is never the case in practice. This may explain why the CRPSS w.r.t. the raw ensemble of this area is highly negative for DRU. The CRPS of QRF (and TQRF) has been computed using $107$ quantiles, rendering perfect prediction of dry events harder and resulting in a deterioration in terms of CRPS over the aforementioned area (see Fig.~\ref{fig:crpss_qrf_raw}). 

The other smaller areas of negative CRPSS w.r.t. the raw ensemble for DRU seem to be caused by numerical instabilities. For example, Figure~\ref{fig:example_forecast_error} presents a numerical instability for a U-Net+GTCND forecast valid on November 3, 2022 at 12:00UTC. It corresponds to heavy precipitation over the Easter part of the region of interest (see Fig.~\ref{fig:example_forecast_obs}). Both raw ensemble and U-Net+GTCND seem not able to correctly predict heavy precipitation, as reflected in the high values of their CRPS (see Fig.~\ref{fig:example_forecast_CRPS_raw} and \ref{fig:example_forecast_CRPS_unet+gtcnd}). However, the CRPS of U-Net+GTCND presents an additional area of high CRPS that is caused by the prediction of precipitation where no precipitation has been observed. This incorrect prediction is characterized by a low value of $L$ (i.e., low probability of dry event), a positive value of $\mu$ and a very high value of $\sigma$ (see Fig.~\ref{fig:example_forecast_L}, \ref{fig:example_forecast_mu} and \ref{fig:example_forecast_sigma}). The abnormally large value of $\sigma$ seems to be caused by a numerical instability and gives a larger probability to large precipitation. The high CRPS over this region associated with a low value of CRPS for raw ensemble causes the CRPSS for U-Net+GTCND w.r.t. the raw ensemble over the test set to be negative (see Fig.~\ref{fig:crpss_unet+gtcnd_raw}).\\

DRUs are able to reach a predictive performance slightly lower but comparable to the QRF. U-Net+CSGD has a slightly better expected CRPS than U-Net+GTCND. In order to be deemed worthy postprocessing methods, U-Net+GTCND and U-Net+CSGD need to be calibrated.

\subsection{Calibration}

Since the ideal forecast (i.e., the true conditional distribution) is unknown, it is impossible to know if a postprocessed forecast has reached the minimum expected CRPS. In order to decompose the contribution of calibration and sharpness to scoring rules \citep{Winkler1977, Winkler1996}, rank histograms are used to evaluate the calibration of the different postprocessing techniques. 

Multiple definitions of calibration exist with different levels of hypotheses (see, e.g., \citealt{Tsyplakov2013, Tsyplakov2020}). The most used definition is probabilistic calibration which, broadly speaking, consists of computing the rank of observations among samples of the forecast and checking for uniformity with respect to observations. If the forecast is calibrated, observations should not be distinguishable from forecast samples, and thus, the distribution of their ranks should be uniform, leading to a flat histogram. The shape of the rank histogram gives information about the type of (potential) miscalibration: a triangular-shaped histogram suggests that the probabilistic forecast has a systematic bias, a $\cup$-shaped histogram suggests that the probabilistic forecast is underdispersed and a $\cap$-shaped histogram suggests that the probabilistic forecast is overdispersed. \cite{Jolliffe2008} proposed a statistical test to assess the uniformity (i.e., flatness) of rank histograms. Moreover, slopes in the rank histograms can be accounted for. \cite{Zamo2016} proposed a test accounting for the presence of a wave in rank histograms. This test is called the Jolliffe-Primo-Zamo (JPZ) test in the following.

\begin{figure}[!ht]
    \centering
    \includegraphics[width=.95\textwidth]{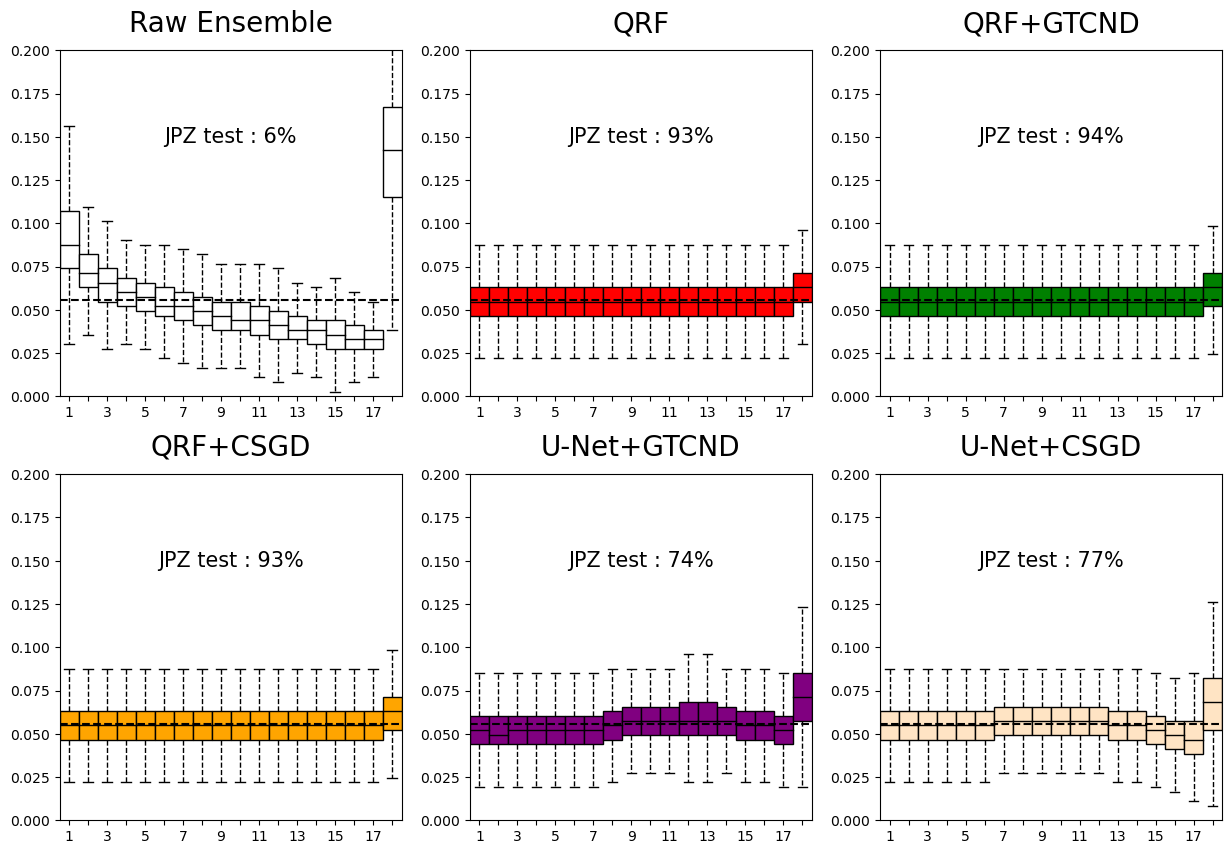}
    \caption{Rank histogram for raw ensemble, QRF, TQRF (namely, QRF+GTCND and QRF+CSGD) and distributional regression U-Nets associated with the GTCND and the CSGD. The hyperparameters are selected as the best performing by cross-validation on the training dataset.}
    \label{fig:rank-histogram}
\end{figure}

\begin{figure}[!h]
    \centering
    \includegraphics[width=\textwidth]{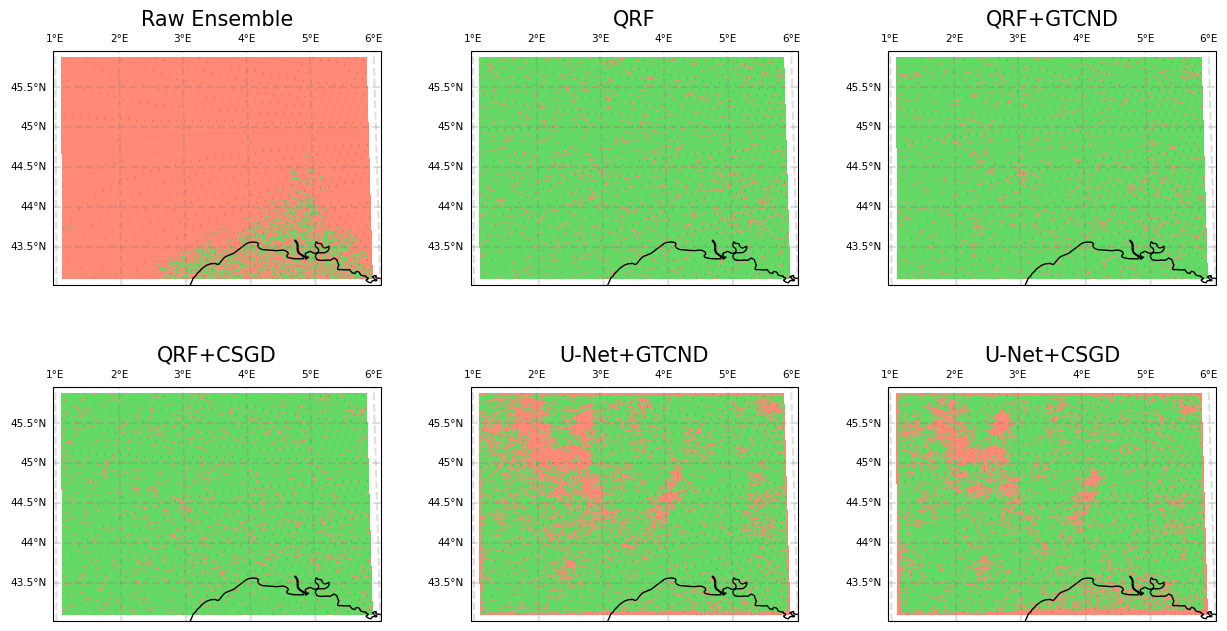}
    \caption{Map of rejection (red) and non-rejection (green) of the flatness of the rank histogram for the forecasting methods considered: raw ensemble, QRF, QRF+GTCND, QRF+CSGD, U-Net+GTCND and U-Net+CSGD.}
    \label{fig:calibration-map}
\end{figure}

To conciliate with the AROME-EPS raw ensemble composed of 17 members, the rank histograms can take 18 different classes and 107 quantiles of the forecasts were produced for the QRF, TQRF and DRU methods (each group of 6 consecutive ranks are gathered as a single rank).

Figure~\ref{fig:rank-histogram} shows the rank histograms of each forecast over the whole grid and the JPZ tests for flatness of rank histograms. As is often the case, the raw ensemble is biased and underdispersed, which is visible by the triangular shape of the rank histograms and the fact that the lowest and highest ranks are over-represented. Its JPZ test confirms that the raw ensemble forecast is not calibrated (only 6\% of grid points do not reject the flatness of the rank histogram). QRF, QRF+GTCND and QRF+CSGD all show very high calibration with JPZ tests not rejecting flatness at 93\%, 94\% and 93\% of grid points. Contrary to what was observed in \cite{Taillardat2019}, no noticeable difference in calibration seems to be present between the QRF and its tail extension. This may be caused by the operational refinement used in the implementation, the fact that different parametric distributions are used and the smaller precipitation accumulations compared to the original article (i.e., 3-h vs. 6-h). DRUs present a lower calibration level compared to QRF-based methods, but their calibration is still significant. The JPZ tests do not reject the flatness hypothesis at 74\% and 77\% of the grid points for the U-Net+GTCND and U-Net+CSGD, respectively. Both DRU forecasts present a slight underdispersion in the right tail revealed by the higher representation of the largest rank in their histograms.\\

Figure~\ref{fig:calibration-map} shows a map of the rejection and non-rejection of the flatness of the rank histogram given by JPZ tests. Calibrated grid points for the raw ensemble are sparsely located over the Mediterranean Sea, the coast and the South of the Rhône valley. QRF, QRF+GTCND and QRF+CSGD are able to calibrate the marginals homogeneously across the region of interest. The areas explaining the lower rate of calibrated grid point for DRU compared to QRF-based methods correspond to high climatological precipitation (see Fig.~\ref{fig:total_precip_test}). The lack of calibration over these areas may be caused by the small depth of the training/validation data (only 3 years) resulting in not enough high precipitation observed. Moreover, the lower performance due to border effects affects the calibration of the DRU forecasts. DRU leads to spatially inconsistent forecasts in terms of calibration whereas the QRF-based methods are homogeneously calibrated over the whole domain. 

\subsection{Extreme events}

\begin{figure}[ht]
    \centering
    \includegraphics[width=.9\textwidth]{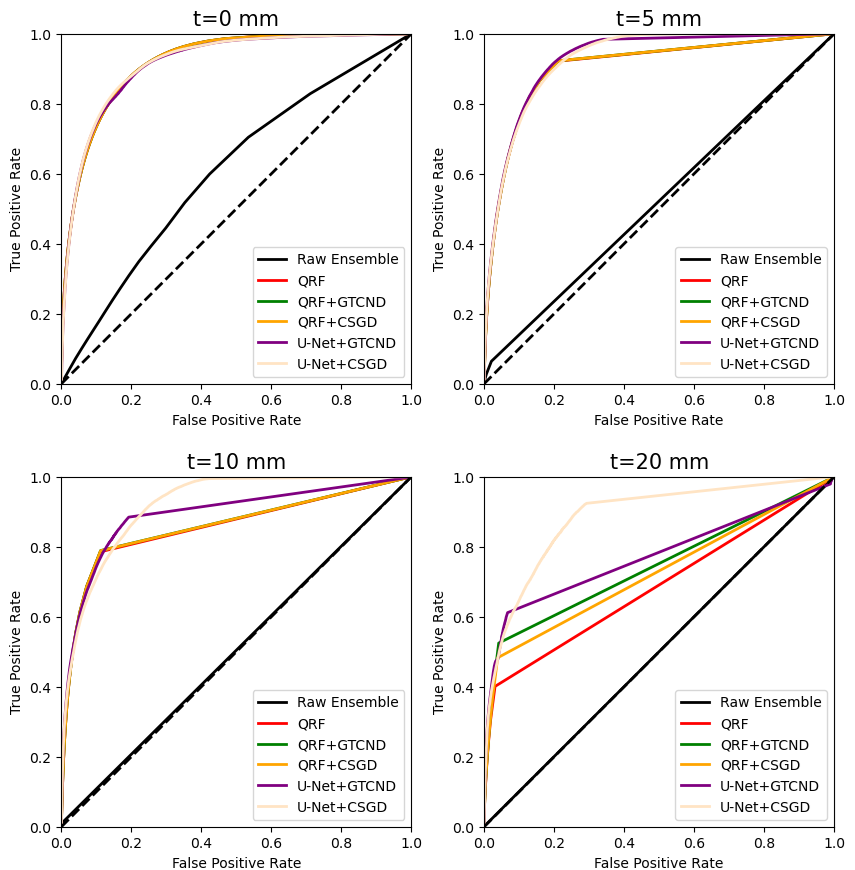}
    \caption{Receiver operating characteristic (ROC) curves of binary events corresponding to the exceedance of a threshold $t\in\{0,5,10,20\}$ (in mm of precipitation). As for Figure~\ref{fig:rank-histogram}, the hyperparameters are selected as the best performing by cross-validation on the training dataset.}
    \label{fig:roc}
\end{figure}

Extreme events are of particular interest. They may lead to the highest socio-economic impacts. However, if verification were to focus only on cases of extreme events, forecasters might be encouraged to propose forecasts that are overly alarming and, thus, of lower general predictive performance. \cite{Lerch2017} pinpointed this phenomenon and named it the \textit{forecaster's dilemma}. Since we have compared the general predictive performance of postprocessing techniques, we can conduct verification focused on extreme events and not be affected by the forecaster's dilemma. 

To focus on forecasts' predictive performance regarding extreme events, we are interested in predicting binary events in the form of the exceedance of a high threshold $t$. We use ROC (receiver operating characteristic) curves to evaluate the discriminant power of forecasts in terms of binary decisions. In particular, ROC curves can inform on the risk of missing an extreme event. Given the binary event $\mathds{1}_{y>t}$ (i.e., exceedance of the threshold $t$), the ROC curve is the plot of the rate of predicted events (i.e., true positive), also called hit rate, versus the rate of false alarms (i.e., false positive). A good forecast should maximize the rate of events detected while minimizing false alarms. In practice, the compromise between the highest hit rate of the method and its lowest false alarm rate depends on the application. In the case of high-impact events, forecasts with a non-negligible false alarm rate may be tolerated if it is accompanied by a better hit rate. In addition to thresholds associated with extreme events, lower thresholds corresponding to lower precipitation events are investigated. Note that grid point by grid point computation of ROC curves does not prevent potential double-penalty effects \citep{Ebert2008}.\\ 

In Figure~\ref{fig:roc}, ROC curves for the exceedance of various thresholds are represented for the raw ensemble, QRF, TQRF and DRU. The lowest threshold is $t=0$mm, which characterizes the prediction of dry events (i.e., absence of precipitation). Raw ensemble has a poor performance regarding the prediction of the presence of precipitation. All the postprocessing methods have comparable performances, as seen in the overlap of their ROC curves. During the cross-validation over the training/validation dataset, the raw ensemble had a better predictive power regarding the prediction of dry events but was still lower than the postprocessing methods (not shown). The threshold $t=5$~mm corresponds to intermediate precipitations. The performance of the raw ensemble already decreases and a difference between DRU and QRF-based methods appears. DRUs have a slightly higher predictive performance compared to QRF-based methods. The raw ensemble lacks resolution because of the nature of its miscalibration (i.e., bias and underdispersion). 

For the highest thresholds $t=10$~mm and $t=20$~mm (corresponding to the quantile of level 0.995 and 0.999, respectively, of the climatology over the region of interest), the ROC curves of the different postprocessing methods can be distinguished. For $t=10$~mm, the performance of the raw ensemble continues to deteriorate and is close to the random guess (dashed line). All the postprocessing techniques are able to maintain a good predictive power but start to noticeably lack resolution, which can be seen in the sudden change of slope. U-Net+GTCND and U-Net+CSGD have a better performance compared to QRF-based techniques which continue to have overlapping ROC curves. U-Net+CSGD has the overall best performance. For $t=20$~mm, the raw ensemble has a performance indistinguishable from a random guess. DRUs are better than QRF-based methods. QRF+GTCND and QRF+CSGD denote from QRF as the tail extension improves predictive performance. QRF+GTCND seems to have a slightly better performance than QRF+CSGD. The gap in performance between U-Net+CSGD and U-Net+GTCND continues to grow and U-Net+CSGD clearly has to the best predictive power w.r.t. the exceedance of the threshold $t=20$~mm.\\

All postprocessing methods compete favorably with the raw ensemble, which has the same predictive performance as a random guess for the highest thresholds ($t=10$~mm and $t=20$~mm). All postprocessing methods have comparable predictive performances for dry events. For heavy precipitation events corresponding to quantiles of levels 0.995 and 0.999, DRUs, and in particular U-Net+CSGD, have a distinctly better predictive power. Moreover, as already observed in \cite{Taillardat2016}, TQRF is able to improve the prediction of heavy precipitation with respect to QRF (even for a light-tailed extension as the GTCND).

\section{Discussion}\label{section:discussion}

We proposed a U-Net-based method, namely distributional regression U-Nets, to postprocess marginal distributions for gridded precipitation data. This approach extends DRN to gridded data by substituting the fully connected NN and embedding module for a U-Net architecture aware of the gridded structure of the data. Simultaneously predicting marginal distributions at each grid point using information from nearby grid points represents a means to account for dependencies between grid points. Both U-Net+GTCND and U-Net+CSGD have predictive performances comparable to the QRF and TQRF in terms of CRPS. DRUs are (probabilistically) calibrated over a large part of the domain studied except for areas associated with the highest precipitation over the test set (see Fig.~\ref{fig:total_precip_test}). This may result from the relatively small training/validation set and could improve with a larger training/validation set. Future studies could try to limit this by emphasizing the learning of high precipitation events using weighted scoring rules for inference. In terms of heavy precipitation, U-Net+CSGD outperforms QRF-based methods.

One of the challenges of the dataset used is the small amount of available training data. This is encountered in practice where consistent data is required, but large reforecast and reanalysis are too computationally expensive. In a more general context, the lack of consistency can be induced at larger time scales by climate change or in specific regions of the world by El Ni\~{n}o forcing.\\

We focused on distributional regression U-Nets where outputs are distribution parameters based on CRPS minimization. DRU can rely on the minimization of other (strictly) proper scoring rules. Moreover, DRU can directly be extended to learn nonparametric distributions such as BQN \citep{Bremnes2020} where the quantile function is a combination of Bernstein polynomials or as HEN (e.g., \citealt{Scheuerer2020}) where the pdf is modeled by the probability of bins.

As U-Net architecture is aware of the spatial gridded structure of the data, specific architectures can also be used for common data structures. We present architectures related to temporal and graph-based structures that are currently used in probabilistic forecasting settings. Their application to postprocessing provides an interesting for future works. For example, if the temporal structure of the data is of interest, recurrent neural networks can be used to predict a parametric distribution. \cite{Pasche2024} proposed to forecast flood risk using high-quantile prediction based on fitting a generalized Pareto distribution via logarithmic score (i.e., negative log-likelihood) minimization. In the case of spatial structure relying on an irregular or more abstract grid (e.g., station network), graph neural networks (GNNs) are able to predict graph-based quantities \citep{Battaglia2018}. \cite{Cisneros2024} used graph convolutional neural networks to learn the parameters of a mixture of a logistic distribution and EGPD via logarithm score minimization to predict wildfire spread. Using the 3D spatial graph-based structures, GNNs are already able to produce deterministic forecasts reaching performance comparable to ECMWF deterministic high-resolution forecasts in performance \citep{Keisler2022, Pathak2022, Bi2023, Lam2022, Chen2023}.

\section*{Acknowledgments}

The authors acknowledge the support of the French Agence Nationale de la Recherche (ANR) under reference ANR-20-CE40-0025-01 (T-REX project) and the Energy-oriented Centre of Excellence II (EoCoE-II), Grant Agreement 824158, funded within the Horizon2020 framework of the European Union. Part of this work was also supported by the ExtremesLearning grant from 80 PRIME CNRS-INSU and this study has received funding from Agence Nationale de la Recherche - France 2030 as part of the PEPR TRACCS program under grant number ANR-22-EXTR-0005 and the ANR EXSTA.

\bibliography{references_unet.bib}

\appendix

\section{Generalized Truncated/Censored Normal Distribution}\label{appendix:gtcnd}

We recall quantities related to the generalized truncated/censored normal distribution (GTCND). Denote $l$ and $u$ the lower and upper boundaries, $L$ and $U$ are the point masses at these boundaries. Since we are working with precipitation, we are interested in the case where $u=\infty$ (implying that $U=0$) and $l=0$, leaving $L$ a parameter to determine along $\mu$ and $\sigma$. Formulas for the general case are available in \cite{Jordan2019}.

The cumulative distribution function (cdf) of the GTCND is
\[
    F_{L,\mu,\sigma}^{gtcnd}(z) = \begin{cases}
        \cfrac{1-L}{1-\Phi(-\mu/\sigma)} \big(\Phi(\frac{z-\mu}{\sigma})-\Phi(-\mu/\sigma)\big)+L &\text{if }z\geq0\\
        0 &\text{if }z<0
    \end{cases}
\]
where $\Phi$ is the cdf of the standard normal distribution. Its quantile function is expressed as
\[
    {F_{L,\mu,\sigma}^\mathrm{gtcnd}}^{-1}(p) = \begin{cases}
        0 &\text{if }p\leq L\\
        \mu+\sigma\Phi^{-1}\left(\frac{(p-L)(1-\Phi(-\mu/\sigma)}{1-L}+\Phi(-\mu/\sigma)\right) &\text{if }p>L
    \end{cases}
\]
for $p\in(0,1)$. The special case of GTCND used here can be expressed using the truncated normal distribution :
\[
    F_{L,\mu,\sigma}^\mathrm{gtcnd}(z) = L \mathds{1}_{z\geq0} + (1-L) N^0_{\mu,\sigma}(z),
\]
where $N^0_{\mu,\sigma}$ is the cdf of the zero-truncated normal distribution.

\subsection*{Moments methods}

\[
    \bbE[\mathds{1}_{X=0}] = L
\]
\[
    \bbE[X] = \mu + \frac{\phi(-\mu/\sigma)\sigma}{1-\Phi(-\mu/\sigma)}
\]
\[
    \Var[X] = \bbE[X^2]-\bbE[X]^2 = \sigma^2 \left\{1 - \frac{\mu}{\sigma}\frac{\phi(\mu/\sigma)}{1-\Phi(-\mu/\sigma)}-\left(\frac{\phi(-\mu/\sigma)}{1-\Phi(-\mu/\sigma)}\right)^2\right\}
\]

\subsection*{Continuous Ranked Probability Score}

\begin{align*}
    \CRPS(F_{L,\mu,\sigma}^\mathrm{gtcnd},y) &=  |y-y_+| + \mu L^2\\
    &+\cfrac{1-L}{1-\Phi(-\frac{\mu}{\sigma})} (y_+-\mu)\left\{2\Phi\left(\frac{y_+-\mu}{\sigma}\right)-\cfrac{1-2L+\Phi\left(-\frac{\mu}{\sigma}\right)}{1-L}\right\}\\
    &+2\sigma \cfrac{1-L}{1-\Phi(-\frac{\mu}{\sigma})} \left(\phi\left(\frac{y_+-\mu}{\sigma}\right)-\phi\left(-\frac{\mu}{\sigma}\right)L\right)\\
    &-\left(\cfrac{1-L}{1-\Phi(-\frac{\mu}{\sigma})}\right)^2 \frac{\sigma}{\sqrt{\pi}} \Phi\left(\frac{\mu\sqrt{2}}{\sigma}\right)
\end{align*}
with $y_+=\max(0,y)$ and $\phi$ the probability density function of the standard normal distribution.

\section{Censored-Shifted Gamma Distribution}\label{appendix:csgd}

We recall quantities related to the censored-shifted gamma distribution (CSGD). The expressions can be found in \cite{Scheuerer2015csgd} and \cite{Baran2016}. The cumulative distribution function (cdf) of the CSGD is
\begin{equation*}
        F_{k,\theta,\delta}^\mathrm{csgd}(z) = \begin{cases}
        G_{k}(\frac{z-\delta}{\theta})\ &\text{if }z\geq0\\
        0\ &\text{if }z<0\\
    \end{cases},
\end{equation*}
with $G_k$ the cdf of the gamma distribution of shape $k$. Its quantile function is expressed as
\begin{equation*}
    {F_{k,\theta,\delta}^\mathrm{csgd}}^{-1}(p) = \delta+\theta\gamma^{-1}(k,p\Gamma(k)),
\end{equation*}
where $\gamma$ is the lower incomplete gamma function, $\Gamma$ is the gamma function and $p\in(0,1)$.

\subsection*{Moments method}
Let $\Tilde{c} = -\delta/\theta$.
\[
    \bbE[X] = (1-G_k(\Tilde{c}))\Big\{\theta k (1-G_{k+1}(\Tilde{c}))-\delta(1-G_k((\Tilde{c}))\Big\}
\]

\begin{align*}
    \bbE[X^2] = (1-G_{k}((\Tilde{c}))\Big\{&k(k+1)\theta^2 (1-G_{k+2}(\Tilde{c}))\\
    &-2\delta k\theta(1-G_{k+1}(\Tilde{c}))\\
    &+\delta^2(1-G_{k}(\Tilde{c}))\Big\}
\end{align*}

\begin{align*}
    \bbE[X^3] = (1-G_{k}(\Tilde{c})\Big\{&k(k+1)(k+2)\theta^3(1-G_{k+3}(\Tilde{c}))\\
    &-3\delta k(k+1)\theta^2 (1-G_{k+2}(\Tilde{c}))\\
    &+3\delta^2 k\theta(1-G_{k+1}(\Tilde{c}))\\
    &-\delta^3(1-G_{k}(\Tilde{c}))\Big\}
\end{align*}

\subsection*{Continuous Ranked Probability Score}

The continuous ranked probability score (CRPS) of the CSGD is
\begin{align*}
    \CRPS(F_{k,\theta,\delta}^\mathrm{csgd},y) &= \theta\Big\{ \Tilde{y}\left(2G_k(\Tilde{y})-1\right) -\Tilde{c} G_k^2(\Tilde{c}) +\theta k\left(1+2G_k(\Tilde{c})G_{k+1}(\Tilde{c})-G_{k}^2(\Tilde{c})-2G_{k+1}(\Tilde{y})\right)\\
    &\ \ \  - \frac{\theta k}{\pi} B(1/2,k+1/2)\left(1-G_{2k}(2\Tilde{c})\right)\Big\},
\end{align*}
where $\Tilde{y}=\frac{y-\delta}{\theta}$, $\Tilde{c}=-\delta/\theta$ and $B$ is the beta function.

\end{document}